\documentclass[10pt,journal]{IEEEtran}
\usepackage{amsmath,amsfonts}
\usepackage{algorithmic}
\usepackage{algorithm}
\usepackage{array}
\usepackage{subfigure}
\usepackage{textcomp}
\usepackage{stfloats}
\usepackage{url}
\usepackage[colorlinks,linkcolor=blue,citecolor=blue]{hyperref}
\usepackage{verbatim}
\usepackage{graphicx}
\usepackage{animate}
\usepackage{graphicx}
\usepackage{float}
\usepackage{amssymb}
\usepackage{booktabs}
\usepackage{multirow}
\usepackage{enumerate}
\usepackage{comment}
\usepackage{tabularx}
\usepackage{diagbox}
\usepackage{color}

\def\BibTeX{{\rm B\kern-.05em{\sc i\kern-.025em b}\kern-.08em
    T\kern-.1667em\lower.7ex\hbox{E}\kern-.125emX}}
\usepackage{balance}
\begin{document}
\title{StableFace: Analyzing and Improving Motion Stability for Talking Face Generation}

\author{
Jun Ling*,
Xu Tan,
Liyang Chen,
Runnan Li,
Yuchao Zhang,
Sheng Zhao,
Li Song,~\IEEEmembership{Senior Member,~IEEE}

\thanks{Jun Ling is with Institute of Image Communication and Network Engineering, Shanghai Jiao Tong University, Shanghai 200240, China. E-mail: lingjun@sjtu.edu.cn. }
\thanks{Xu Tan is with Microsoft Research Asia (MSRA), Beijing, 100080, China. E-mail: xuta@microsoft.com.}
\thanks{Liyang Chen is with Tsinghua University, Shenzhen, 518055, China. Email: cly21@mails.tsinghua.edu.cn.}% <-this % stops a space
\thanks{Runnan Li, Yuchao Zhang, and Sheng Zhao are with Microsoft Azure Speech, Beijing, 100080, China, E-mail: \{Runnan.Li, Yuchao.Zhang, Sheng.Zhao\}@microsoft.com}.
\thanks{Li Song (Corresponding author) is with the Institute of Image Communication and Network Engineering, Shanghai Jiao Tong University, Shanghai 200240, China, and also with the MoE Key Laboratory of Artificial Intelligence, AI Institute, Shanghai Jiao Tong University, Shanghai 200240, China. E-mail: song\_li@sjtu.edu.cn.}
\thanks{*Part of this work was conducted at Microsoft.}
}

\markboth{Journal of \LaTeX\ Class Files,~Vol.~18, No.~9, July~2022}%
{How to Use the IEEEtran \LaTeX \ Templates}

\maketitle

\begin{abstract}

While previous speech-driven talking face generation methods have made significant progress in improving the visual quality and lip-sync quality of the synthesized videos, they pay less attention to lip motion jitters which greatly undermine the realness of talking face videos. What causes motion jitters, and how to mitigate the problem? In this paper, we conduct systematic analyses on the motion jittering problem based on a state-of-the-art pipeline that uses 3D face representations to bridge the input audio and output video, and improve the motion stability with a series of effective designs. 
We find that several issues can lead to jitters in synthesized talking face video: 1) jitters from the input 3D face representations; 2) training-inference mismatch; 3) lack of dependency modeling among video frames. 
Accordingly, we propose three effective solutions to address this issue: 1) we propose a gaussian-based adaptive smoothing module to smooth the 3D face representations to eliminate jitters in the input; 2) we add augmented erosions on the input data of the neural renderer in training to simulate the distortion in inference to reduce mismatch; 3) we develop an audio-fused transformer generator to model dependency among video frames. Besides, considering there is no off-the-shelf metric for measuring motion jitters in talking face video, we devise an objective metric (Motion Stability Index, MSI), to quantitatively measure the motion jitters by calculating the reciprocal of variance acceleration. Extensive experimental results show the superiority of our method on motion-stable face video generation, with better quality than previous systems\footnote{Video samples: \url{http://stable-face.github.io}.}. 
\end{abstract}


\begin{figure}
  \centering
  \includegraphics[width=1\linewidth]{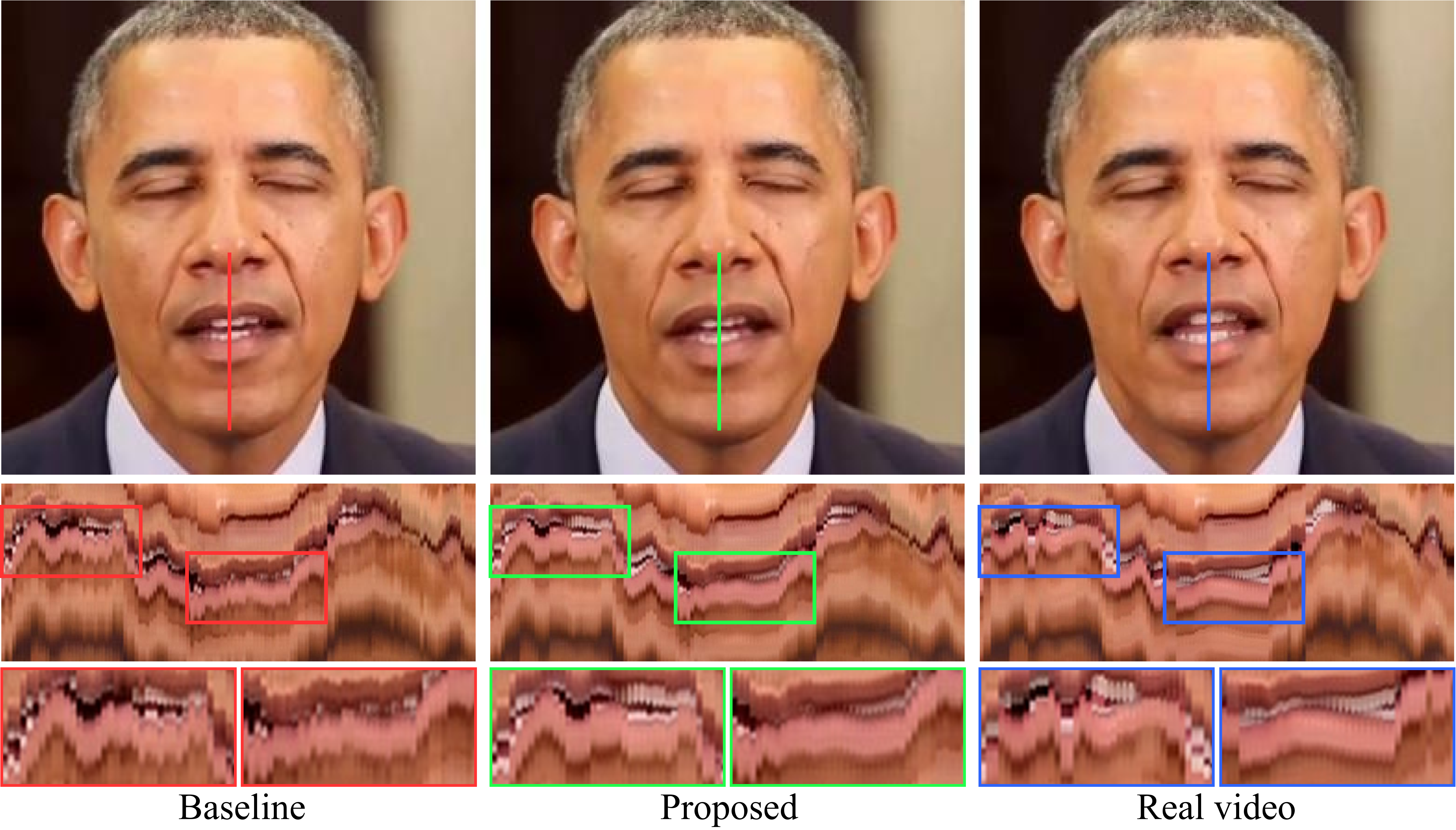}
  \caption{Illustrations of videos with/without motion jitters. We generate talking face videos with the baseline (\emph{top left}) and our proposed method (\emph{top middle}), the audio is extracted from the real video (\emph{top right}). For each video, we concatenate the vertical slice (the vertical red/green/blue line) in each frame along the time, and then show the results at the bottom of each video. It can be observed that the baseline figure has more jittering patterns than the proposed and real videos. We highly recommend readers watch our demo videos for better experience. 
  }
  \label{fig:jitter_vis_comp_metric}
\end{figure}

\section{Introduction}
\label{sec:intro}
Talking face video generation~\cite{chen2019hierarchical,zhou2020makelttalk,prajwal2020lip,vougioukas2020realistic,zhou2021pose,wu2021imitating,chen2020duallip,chen2020talking,yi2020audio,ji2021audio} shows promising potential for multimedia applications, such as filmmaking, cartoon, TV shows, newscasting, virtual assistant, and vitual avatar in metaverse, and \emph{etc}. A talking face synthesizing system usually takes a speech sequence as input\footnote{There are also some works~\cite{fried2019text,yao2021iterative,chen2020duallip} taking text as input. However, we mainly consider speech as input here due to several reasons: 1) Even if using text as input, we still need to synthesize speech from text since a talking face always needs speech; 2) We need to predict the duration of text to align with taking video if using text as input, while speech can be easily aligned with talking video.} and generates a sequence of realistic talking face images that conveys the visual contents of the speech.

Generating realistic talking face video is valuable but also challenging in that it not only requires high image quality and lip-sync quality but also good motion stability. For example, as shown in Fig.~\ref{fig:jitter_vis_comp_metric}, the motion jitters (the irregular or unnatural movements~\cite{seshadrinathan2009motion} of the mouth or head in several continuous frames) severely undermine the realness of talking face videos. 
Although various practical talking face generation approaches have been proposed in recent years, most of these studies tends to focus on improving the visual quality of each synthesized image\cite{ji2021audio,thies2020neural,guo2021adnerf,lu2021live}, or the lip-sync quality~\cite{lahiri2021lipsync3d,vougioukas2020realistic,prajwal2020lip}. Meanwhile, these works usually quantitatively evaluate the video quality with single-frame-based metrics (PSNR, SSIM~\cite{wang2004image}, LMD~\cite{chen2018lip}) and measuring video realness via human-centered subjective experiments, paying less attention to the motion jittering problem, thus leaving the challenge unsolved.

In this paper, we delve into the motion jittering problem with systematic analyses and then mitigate the problem with effective solutions. Specifically, we choose a basic talking face generation pipeline as the baseline model, because this method is very representative and has been widely used in previous works~\cite{thies2020neural,yi2020audio,lahiri2021lipsync3d,song2021tacr,wen2020photorealistic,zhang2021facial}. In this pipeline, we first estimate the mouth-related expression parameters from the extracted audio features. Then, the expression parameters, as well as the shape and pose parameters (from the background images) are combined as the input of a 3D face model~\cite{feng2021deca} to render animated face shapes (3D face representations used in this work). To synthesize target realistic face images, we utilize a neural renderer to predict the images given the concatenation input of the background images and animated face shapes. However, using 3D face representations is inefficient to describe facial details due to the information of the tongue and teeth has been lost. To compensate the information loss, we design an Audio Fusion Module (shown in Fig.~\ref{fig:detailed_components} (\emph{top left})) that fusion audio features with the image features.

With some preliminary analyses, we find several key reasons that incur the motion jitters: 
\textbf{(1)} First, jitters from 3D face representations. The 3D face representations provide the detailed mouth information and head poses, and serve as the input of neural renderer. Since they are extracted by the 3D face model on a single image without considering the information from adjacent frames, they are not smooth across frames and can have jitters, and will provide the unstable ground-truth labels for audio-to-expression prediction network. 
\textbf{(2)} Second, training-inference mismatch in neural renderer. In the training phase, the neural renderer is optimized to generate a realistic image from the background image and the face shape. However, the background image and the face shape are from the same target image, which can relieve the optimizing procedure but cause mismatch in inference. In inference, the face shape where the mouth area has been changed due to the new audio, which does not match the same background. When concatenating together, the model needs to generate a realistic face with mouth part from the face shape while the rest from the background image. This makes it difficult for the model to handle this mismatch that has never been seen in training, thus imposing uncertainty and causing motion jitters in the rendered images. 
\textbf{(3)} Third, the lack of dependency modeling across consecutive frames in neural renderer. Neural renderer in current framework~\cite{thies2020neural} learns to synthesize each image independently without modeling the dependency across consecutive frames, thus failing to generate motion stability talking face videos.

To address these issues that cause motion jitters, we propose several effective solutions. \textbf{(1)} First, we propose to remove the jitters from the 3D face representations. A simple way for smoothing is to use simple moving average or manually designed smoothing weights. However, both of them cannot handle the mouth movements with varying speed (fast or slow), either incur the problem of over-stable motions and eliminate the differences between similar pronunciations, or produce less motion stable results. To combat this, we train a smoothing weight estimation network to adaptively predict different weights for each frame given the 3D face expressions. 
\textbf{(2)} Second, we introduce augmented erosion to background images in training to simulate the mismatch in inference. The augmented erosion module randomly eroded the mouth regions with different shape images to simulate the potential mismatch in inference.  With augmented erosion, our neural renderer is more robust to the distortions in the mouth regions and reduce jitters. 
\textbf{(3)} Third, treating talking face generation as a sequence-to-sequence generation task, we develop a transformer-based dependency modeling module, and embed it into the neural renderer. Our dependency module takes the advantages of transformer in temporal relations modeling, making contributions to improve motion stability.

Equally to the study of motion jittering problem, quantitatively evaluating the motion jitters is also significative for talkindg face generation which not only reduces the cost that subjective experiments take but also evaluates the motion stability on videos without introducing individual user bias. However, it is difficult to evaluate motion jitters in that motion jitters cannot be observed from a single image or two but a sequence of images, which makes the quality of motion stability hard to be defined and measured. The absence of off-the-shelf metrics which can measure the motion stability or motion jitters in talking face videos, impedes further exploration of motion-stable talking face generation. 
To bridge this gap and facilitate future research, we devise an objective metric, namely Motion Stability Index (MSI), to measure motion stability in talking face video. In particular, we utilize the reciprocal of the variance of accelerations of each key point in the face video to measure the motion stability. Experiments show that the Pearson correlation coefficients between MSI and subjective scores on motion stability reach to 0.438, which demonstrates the efficacy of our MSI. More details can be found in Sec.~\ref{subsec:motion_jittering_metric}.

To summarize, our main contributions in this paper can be marked as follows: 

(1) We systematically study the motion jittering problem, and analyze the causes of motion jitters. To our best knowledge, this is the first work that focus on motion jittering problem in talking face generation task. 

(2) To address the motion jitter issue, several systematical designs are proposed in framework, including adaptive smoothing module, augmented erosion, and transformer-based dependency modeling module to improve the motion stability of synthesized talking face videos. 

(3) To facilitate the research of talking face generation, we propose an effective objective metric (MSI) to quantitatively evaluate motion stability in face videos. Ablation studies demonstrate the efficacy of proposed metric.

\section{Related Work}
\label{sec:backgrounds}

\subsection{Talking Face Generation}
\label{subsec:talking_face_generation}

\noindent
\textbf{Video-Driven vs Audio-Driven. }
Existing works on talking face generation can be divided into three categories: video-driven, audio-driven, and text-driven methods, according to the driving source of facial animation. 

Video-driven talking face generation aims to reenact the existing facial image with external driving source, \emph{e.g.}, another talking image or video~\cite{thies2016face2face,ling2020toward,wiles2018x2face,kim2018deep,nagano2018pagan,xue2020realistic,tang2022generative}. Although these works can synthesize talking face images with good visual quality, video-driven methods require an additional driving video, and still need another lip-sync audio to compose a complete talking face video. Audio-driven talking face generation employs speech as the driving source and synthesizes face video in sync with the input speech content. In this paper, we focus on audio-driven talking face generation, since it has some advantages over its video-driven counterparts: 1) they are more essential to talking face generation since the goal is to generate talking face from talking audio, and 2) they do not need additional driving video which makes them more generally applicable to different scenarios.

\begin{figure*}[!htbp]
  \centering
  \includegraphics[width=0.9\linewidth]{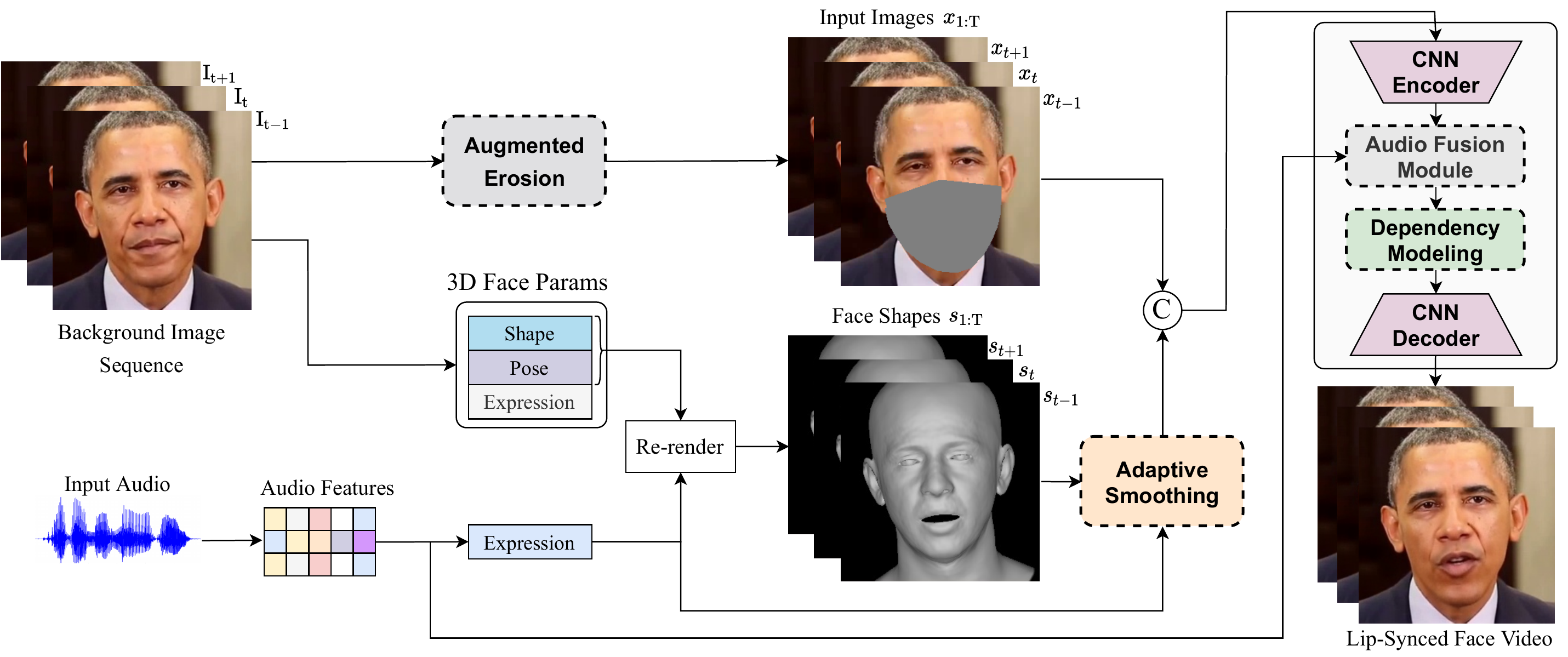}
  \caption{Overview of the pipeline. We first predict the 3D expression parameters from audio features, and then combine the shape, pose and texture parameters from input images with the predicted expressions to render the target face shapes. We propose several effective components to synthesize face images that are not only motion-stable, but are synchronized with input speech. The modules with dashed borders represent our key contributions on the baseline. 
  }
  \label{fig:overall_pipeline}
\end{figure*}

\noindent
\textbf{End-to-End vs Two-Stage.} 
Some works tackle audio-driven talking face generation in an end-to-end way, which directly generate talking images from speech features~\cite{chung2017you,chen2018lip,zhou2019talking,prajwal2020lip,chen2020duallip,zhou2021pose}. The shared feature of these methods is that no intermediate facial geometry utilized to bridge the neural renderer and input speech. However, there are information mismatch between input audio and output video: 1) audio content has strong correlation with the movements around the mouth area while has very weak correlation with the head pose and eye gaze; 2) the static features such as shape or texture of the talking face are not in good correlation with the audio content. 
Another body of works propose two-stage methods that use intermediate representations to represent the target talking face, and first predict the intermediate representations from the audio input, and then synthesize the talking face from the predicted representations using a neural renderer. 
Besides, since audio content has very weak correlation with the head pose and eye gaze, if generating the whole intermediate representations that contain lip movement, expression, head pose, and eye gaze, the model will suffer from one-to-many mapping and ill-posed problem. Thus, some works~\cite{thies2020neural,lahiri2021lipsync3d,song2021tacr,ji2021audio,suwajanakorn2017synthesizing,yi2020audio} propose to only generate the intermediate representations in the mouth area, and then concatenate the mouth representations with background image to generate whole talking image. In this paper, we choose the two-stage methods and predict mouth related representations while directly using the background image from target face, which can lead to better quality in synthesized talking face. Although two-stage method is chosen in our paper, the similar causes (jitters from inputs and lack of dependency modeling) are the same in both two-stage and end-to-end methods.

\noindent
\textbf{2D Face Geometry vs 3D Face Representation. }
The intermediate representations of talking face (as introduced above) can be chosen as 2D face geometry (facial landmarks or face parsing maps)~\cite{siarohin2019first,wang2019few,chen2019hierarchical,zakharov2019few,zakharov2020fast,xue2020realistic,lu2021live} or 3D face representations~\cite{blanz1999morphable,thies2020neural,wen2020photorealistic,feng2021deca,wu2021imitating,doukas2021headgan}. Using 2D face geometry as intermediate representations is simple, and has several drawbacks: 1) it is coarse-grained to represent a talking face with 2D face geometry; 2) it requires large amount of high-quality videos for training; moreover, 3) it easily causes identity leakage and face geometry deformation since incorrectly predicted 2D geometry contain distortions in human face geometry. Recent works usually use 3D face representations as intermediate features, since it can relieve the above issues in 2D face geometry. In this paper, we follow the same setting of 3D face representations~\cite{thies2020neural} for talking face generation, but focus on solving the motion jittering problem.

\subsection{Vision Transformer}
\label{subsec:vision_transformer}
Recently, transformer~\cite{vaswani2017attention} has been becoming the most popular in sequence modeling long-term correlations in various research fields~\cite{devlin2018bert,ren2019fastspeech,fan2022faceformer}. Typically, the transformer consists two main conponents, incorporate a multi-head self attention mechanism over the tokens and a feedforward module. Due to its capability in long-range relationship modeling, transformer-based models have been applied to various kinds of vision tasks, such as visual recognition~\cite{dosovitskiy2020image}, 3D facial animation~\cite{fan2022faceformer,chen2022transformers2a}, inpainting~\cite{liu2021fuseformer,ren2022dlformer}, object detection~\cite{carion2020end,zhu2020deformable}, image synthesis~\cite{zeng2021improving}, image harmonization~\cite{guo2021image} \emph{etc}. 

In the fileds close to the talking face video generation, as far as our knowledge concerns, most of recently proposed methods deal with 3D meshes generation~\cite{fan2022faceformer,li2020learning} or facial blendshape coefficients estimation~\cite{chen2022transformers2a}, while fewer of them focus on photo-realistic talking face generation. Fan \emph{et. al.} propose FaceFormer~\cite{fan2022faceformer} designs an autoregressive transformer-based architecture that predicts moving coordinates for 3D facial mesh animation. Li \emph{et. al.}~\cite{li2020learning} propose a two-stream transformer to model the motion distribution and capture long-term dependency. Chen \emph{et. al.}~\cite{chen2022transformers2a} propose a MOE-based transformer, to generate facial animation coefficients from speechwhich should be fed into graphic engine to generate animations. Despite these methods achieve significant improvements in animation generation, they cannot be adapted to photo-reslistic video generation framework because the input and output data are both 4 dimensions, \emph{i.e.}, $\mathbb{R}^{\rm T\times H\times W\times C}$. Therefore, in this paper, we treat photo-realisitc talking face generation as a frame-based sequence-to-sequence generation task and build a transformer-based architecture to capture dependencies among consecutive video frames.

\subsection{Motion-Stable Video Generation}
\label{subsec:motin_stable_video_generation}
Besides the lip-sync quality and high image fidelity, in the talking face videos, the frame visual consistency (\emph{e.g.}, illumination, or color) and motion jitters also harm its naturalness and realness. Previous works have tried different approaches to improve the quality of synthesized images to make people feel more realistic over each frame or improve the lip-sync quality of the talking videos. To improve frame consistency, some talking face generation methods introduce autoregressive generation strategy~\cite{song2021tacr,lahiri2021lipsync3d} by "looking at previously generated frames" in the neural renderer, or incorporate a temporal discriminator to judge the video quality on multiple consecutive frames~\cite{chen2019hierarchical,vougioukas2020realistic,prajwal2020lip}. However, these methods are not designed for improving motion stability, and are not efficient enough to mitigate the motion jittering problem (which can be clearly observed in their released talking face videos). Yu \emph{et. al.}~\cite{yu2020multimodal} utilize motion flow to warp previous images to improve consistency. While achieving better temporal consistency, the optical flow only represents the motion between two adjacent frames, but not long-range dynamics. Meanwhile, inaccurate and unstable motion flows also hinder stable motion generation. 

Beyond talking face generation task, some works try to solve the temporal inconsistency problems~\cite{lai2018learning,lei2020blind} that are incurred by illumination or color changing across the adjacent frames. Unfortunately, this kind of temporal jitters of appearance in consecutive frames is not the same as motion jitters which care more in the object motions and less in appearance. Some other works try to improve video stability for hand-held captured videos due to the high-frequency camera jitters occur in video capturing procedure~\cite{matsushita2006full,liu2009content,wang2018deep,li2022deep} but not the motion jittering problem in video synthesis.

%-------------------------------------------------------------------------
\section{Method}
\label{sec:method}
In this section, we first introduce the overall setting of baseline model (Sec.~\ref{subsec:overall_pipeline}). Then, we roundly analyze the causes of motion jittering problem (Sec.~\ref{subsec:analysing_motion_jittering}) and propose our solutions (Sec.~\ref{subsec:improving_motion_consistency}). Afterwards, we introduce the proposed objective metric for measuring the motion stability (Sec.~\ref{subsec:motion_jittering_metric}) and the loss functions(Sec.~\ref{subsec:loss_function}).

\begin{figure*}
    \centering
    \includegraphics[width=0.95\linewidth]{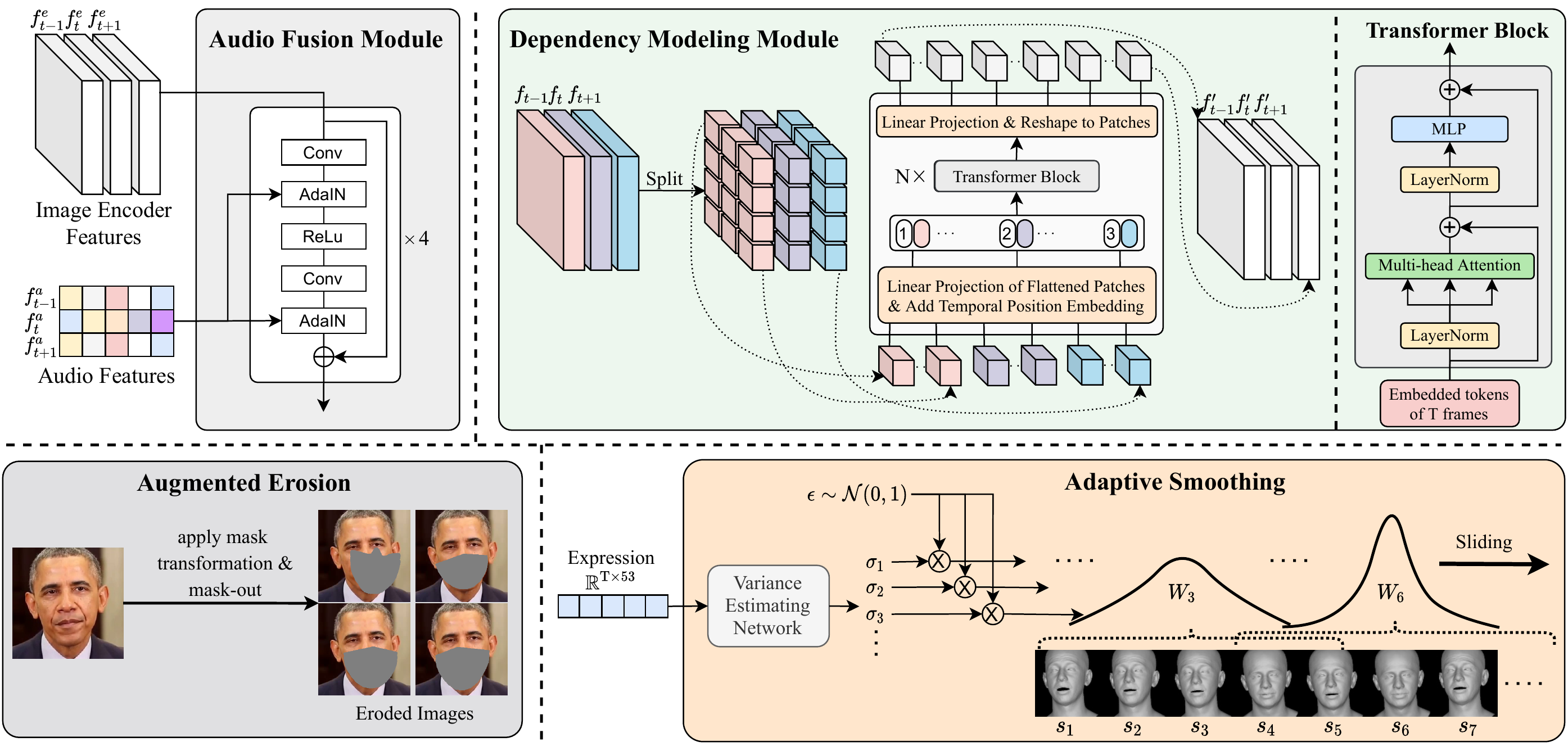}
    \caption{Detailed illustrations of our designs. In our pipeline, we perform Augmented Erosion (\emph{bottom left}) on the input background images and Adaptive Smooth (\emph{bottom right}) the input shapes. A convolutional-based image encoder is employed to extract image encoder features from the concatenated inputs. Then, Audio Fusion Module (\emph{top left}) takes as input the image encoder feature and the audio features, and output the fused feature (denoted by $f_{t-1}, f_{t}, f_{t+1}$). Further, we learn temporal dependencies on the consecutive frames via the Dependency Modeling Module (\emph{top right}). Finally, the output features (denoted by $f'_{t-1}, f'_{t}, f'_{t+1}$) can be decoded as a sequence of images that are both motion stable and lip-synced with the input audio.
    }
    \label{fig:detailed_components}
\end{figure*}

\subsection{Baseline Overview}
\label{subsec:overall_pipeline}
To analyze and mitigate the motion jittering problem, we investigate previous approaches that can handle talking face generation on personalized talking face videos, and then design a baseline. In brief, the baseline is audio-driven, two-stage, and uses 3D face parameters as the intermediate representations between audio input and video output. In this section, we introduce each component of our baseline as follows.

% \paragraph{3D Face Representation}
\noindent
\textbf{3D Face Representations. }
3D face model~\cite{blanz1999morphable,feng2021deca} has been used to disentangle facial shape, texture, expression, and other facial properties. With 3D face model, one can reconstruct the in-the-wild face with one dimensional parameters vector, \emph{i.e.}, shape, expression, texture, and pose parameters. As a typical pipeline, taking 3D face representations to bridge the input audio and output video has been validated by recent competing methods~\cite{thies2020neural,wen2020photorealistic,song2021tacr,ji2021audio,yi2020audio}
. The 3D face model is used in training and inference respectively as follows. In training, we first extract 3D face parameters from the target images b{}y DECA~\cite{feng2021deca}, which include facial expression, shape, pose, and texture. These parameters have two usages: 1) the facial expression of the mouth area is taken as the training target of the audio2expression model in audio processing (which will be introduced later); 2) the expression, shape, and pose parameters of the whole images are used to render the face shapes, which are concatenated with the background images.

\noindent
\textbf{Audio Processing. }
We employ a pre-trained audio feature extraction model~\cite{amodei2016deep} to extract audio features (denoted as $f^{a}_{1:\rm T}\in\mathbb{R}^{\rm T\times 64}$), and then predict the mouth-related 3D expression parameters (denoted as $\beta \in \mathbb{R}^{\rm T\times 53}$, where T equals the number of video frames, and 53 denotes the dimension of 3D face expression parameters in our experiment) from audio features using a robust audio2expression model (Transformer-S2A~\cite{chen2022transformers2a}). Taking a sequence of audio features as input, Transformer-S2A estimates a sequence of 3D expression parameters with several stacked transformer layers. 

% \paragraph{Neural Renderer}
\noindent
\textbf{Neural Renderer. }
The neural renderer in this baseline takes the eroded background images and 3D face representations as input, and synthesizes new images frame-by-frame without modeling dependencies among consecutive frames. The neural renderer consists of three main components: an encoder, a bottleneck, and a decoder. The encoder consists of one convolutional layer (kernel size 7, stride 1) and 2 strided convolution layers (kernel size 4, stride 2), each followed by instance normalization and leaky-relu activation. We use 8 residual blocks as~\cite{he2016deep} for the bottleneck. For the decoder, we use 2 transposed convolution layers (kernel size 4, stride 2) and 1 convolution layer (kernel size 7, stride 1) and tanh activation to output the final images.

\subsection{Analyzing Motion Jittering Problem}
\label{subsec:analysing_motion_jittering}
We consider the generation procedure as a sequence-to-sequence mapping problem ${\rm \hat{I}}_{1:\rm T} = G(s_{1:\rm T}, x_{1:\rm T},{\theta})$, where $s_{1:\rm T}$ and $x_{1:\rm T}$ represent the input 3D face representations and eroded images from time stamp of 1 to $\rm T$, and ${\theta}$ is the network parameters. Since the neural renderer acquires background visual information (hair, backgrounds) from the eroded background images, it is easy for the neural renderer to synthesize the background-related regions. However, the mouth and jaw parts are more challenging to deal with because 1) the synthesized mouths and jaw should match the background images, and 2) the motions of the lip and jaw are complex and diverse due to the diverse content of input audio. With systematic analyses, we find several causes that may incur motion jitters in the synthesized videos, including causes from $s_{1:\rm T}$, $x_{1:\rm T}$, and ${\theta}$.

First, jitters from the 3D face representations $s_{1:\rm T}$. We adopt the choice of 3D face shapes to bridge the neural renderer and audio processing module. However, the facial parameters are extracted by the 3D face model on single image or predicted by audio processing module, they are not stable across consecutive frames and can have jitters. Consequently, the jitters from the input parameters incur the jitters of 3D face shapes, which are then input to the neural renderer. A straightforward solution to alleviate this jitters is to smooth the input of the neural render. Instead of smoothing the 3D parameters in parametric space, we propose to smooth the 3D face shapes in geometry space due to the fact that the 3D face shapes are the direct input of the neural render.

Second, training-inference mismatch. In training, we extract the 3D face parameters from the background images and render the face shapes as the input of neural render. Similar to~\cite{song2021tacr,thies2020neural,wen2020photorealistic}, we mask-out the mouth part and concatenate the eroded image with the face shape in channel-wise manner, and treat them we the input of neural renderer. However, the eroded image and the face shape are from the same target image in training, which relieves the optimizing procedure. In inference, the eroded image is from the same background video, which is not complementary with the new face shape where the mouth area has been changed due to the new audio. Consequently, the neural render should generate a realistic face with mouth part from the face shape while the rest from the eroded image, which makes it difficult for the neural renderer to deal with this distortion and producing more errors around the mouth boundary.

Third, lack of dependency modeling among video frames in the neural renderer. Each image frame in a video is not independent but correlated with its adjacent frames. However, the current neural renderer generates each frame independently, without considering the information from adjacent frames. Therefore, synthesizing each frame without "looking behind and ahead" probably produces motion jitters in face video. To address this issue, we propose to apply a dependency modeling module to enforce dependencies learning and improve the motion stability for the current neural render.

% \paragraph{Adaptive Smoothing. }
% \noindent
\subsection{Improving Motion Stability}
\label{subsec:improving_motion_consistency}

% \noindent
\subsubsection{\textbf{Adaptive Smoothing}}
To relieve the motion jitters caused by the independent extraction of 3D face parameters, we propose to smooth the face shapes among adjacent frames. We define the smoothing weight as $W \in \rm \mathbb{R}^{T\times K}$, where $\rm T$ is the number of frames. For each $t \in [1, \rm T]$, ${W}_t \in \rm \mathbb{R}^{K}$, where $\rm K$ is the smoothing width (\emph{i.e.}, the number of adjacent frames taken for smoothing) and is usually an odd number. Therefore, smoothing over the face shapes can be written as: 
\begin{equation}
    \widetilde{s}_{t} = \sum_{k=- (\rm K-1)/2}^{(\rm{K}-1)/2}{W}_{t, k }*s_{t+k},
\end{equation}
where $s_{t+k}$ is the face shape in the $ (t+k)$-th frame, $\widetilde{s}_{t}$ is the smoothed face shape in the $t$-th frame.

\begin{table}[!htbp]
    \caption{Detailed architecture of adaptive smoothing module. T is the number of samples in image sequence. }
    \centering
    \begin{tabular}{ccc}
    \toprule
      Layer  & Kernel Size & Output Size \\
       \midrule
        Conv1D-BN-ReLU & 3 & (T$\times$32) \\
        Conv1D-BN-ReLU & 3 & (T$\times$32) \\
        Conv1D-BN-ReLU & 3 & (T$\times$16) \\
        Conv1D-BN-ReLU & 3 & (T$\times$16) \\
        Conv1D-Sigmoid & 3 & (T$\times$K) \\
    \bottomrule
    \end{tabular}
    \label{tab:adaptive_smoothing_net}
\end{table}

There are different choices for the smoothing weight $W$: 1) a handcraft and fixed weight; 2) a global (i.e., for any $i, j \in [1, \rm T]$ , $W_i = W_j$) but learnable weight (\emph{i.e.}, $W$ is optimized in an end-to-end way); 3) an adaptive weight (\emph{i.e.}, where $W_i$ can be different from $W_j$ for any $i\neq j$). In our experiments (see Fig.~\ref{fig:adaptive_smoothing_analyses}), we find that using handcraft and fixed weight often leads to over-smoothing that eliminates the subtle expression variations in fast movement or less-smoothing that cannot deal with the motion jitters. For global but learnable weight, it is the learned weight that average on the whole dataset but cannot solve the problem of fixed weight. Thus, we choose to adaptively learn the weight, i.e., each $W_t$ is learned based on the current and adjacent frames of the input face shapes.

\noindent
\subsubsection{\textbf{Augmented Erosion}}
To address the training-inference mismatch problem in the neural renderer for better motion stability, we add augmented erosions on the background images in training to simulate the patterns of mismatch in inference. We employ two ways to add augmented erosions: 1) we add random noise to the original facial expressions on the mouth area, and then create the eroded images via mouth area mask generated by~\cite{feng2021deca}; 2) we randomly erode/dilate and shift/rotate the mask to simulate mismatch patterns in inference phase. The generator benefits from this operation and learns to synthesize images around the mouth region from different image-shape patterns, thus becomes more robust in inference stage. The results of augmented erosion on background image can be viewed in Fig.~\ref{fig:detailed_components}.

\noindent
\subsubsection{\textbf{Transformer-based Dependency Modeling}}
\label{subsec:dependency_modeling}
Similar to natural language or speech, videos can be viewed as sequence data that consists of multiple consecutive frames. In order to eliminate the motion jitters in the synthesized video, dependency modeling among video frames is incorporated into the neural renderer. Recently, transformer-based methods have been introduced into various areas and outperform other RNN-based methods in handling sequential data, eschewing recurrent modeling but merely relying on self-attention mechanism. To generate talking face videos in a sequence-to-sequence generation manner and explicitly model frame dependencies, in this work, we design a transformer-based dependency modeling module, employing several stacked Transformer blocks in the convolutional encoder-decoder architecture. 

As shown in Fig.~\ref{fig:detailed_components} (top right), the Dependency Modeling Module takes as input the fused features $f_{1:\rm T}$, where $f_{i}\in \mathbb{R}^{h\times w \times c}, h=w=64, c=256$. To formulate the sequence embedding for transformer, we first split each fused feature into $p\times p$ smaller patches for each sample $f_{i}$. Then all patch (each patch has shape $64/p\times 64/p \times c$) will be flattened and linearly projected into tokens $\mathbf{Z}\in \mathbb{R}^{(T\cdot p \cdot p)\times d}$, where $d=512$. Similar to ViT~\cite{dosovitskiy2020image}, our transformer encoder is stacked by $N$ transformer blocks, each block consists of two main components: the spatial-temporal self-attention mechanism and the feedforward layers. Taking as input the sequential data of projected patches, the dependency modeling module performs self-attention and feature learning for both temporal and spatial dependencies. The output of our transformer block will be linearly projected and reshaped to $p\times p$ patches, and then be composed and fed into a convolutional decoder, which recovers the facial details from the output features $f'_{1:\rm T}$. In our experiments, we set $p=4$, and $N=4$. 

Benefiting from the transformer-based dependency modeling module, the neural renderer is capable of generating talking face videos with more stable and consistent motion. Compared to the baseline that independently synthesizes each face image, our transformer-based neural renderer is capable of learning inter-frame dependency and synthesizing talking face videos with better motion stability and frame consistency. Compare to RNN-based models, our transformer-based neural render can not only capture long-range and temporal dependencies conditioning on the whole sequence via self-attention mechanism, but also enable both training and inferencing in parallel, which requires less latency than RNN-based methods.

\subsection{Motion Stability Index (MSI)}
\label{subsec:motion_jittering_metric}
% 要先讲why
% 我们为什么要提出这个
% 是因为这个领域很急需，没有很好的替代品
As mentioned in Sec~\ref{sec:intro}, quantitatively evaluating the motion stability of a talking face video is valuable and challenging. However, as far as our knowledge is concerned, there is no available objective metric to measure the the motion stability in previous works~\cite{lahiri2021lipsync3d,thies2020neural,song2021tacr,lu2021live}, thus forcing us to solely rely on subjective user study. To bridge the gap, we resort to develop an effective metric that measure the motion stability in talking face videos.

To mesure the motion stability, we start from the motion jitters in talking face videos. Motion jitters describe the irregular motion variations in a video, and the more stable motion indicates the better smoothness of the motion variations. Based on this analysis, we propose to use the variance of the motion variations to measure motion jitters. For simplicity we represent the motion variations by accelerations. 
To acquire the accelerations, the first thing to do is acquiring the motion in each video frame. Intuitive choice to represent motion in video is using optical flow~\cite{fleet2006optical} which measures the motion of each pixel between two frames. However, optical flow has two drawbacks. 
First, the pixels in optical flow have no accurate semantic labels, thus we cannot distinguish the mouth from eyes or nose via optical flow map. 
Second, optical flow estimates the pixels motion between two consecutive frames, thereby imposing difficulties in forming the trajectory of a certain object of interest, such as the lip. 
Therefore, in this work, we adopt facial key points to represent the key motions of talking faces. With the coordinates of each key point, we can easily calculate the motions and acquire their trajectories. To achieve this, we first track the facial landmarks via face alignment tool~\cite{wood2021fake}, and use them to calculate the motions.

Denote the landmarks at each frame $t$ as $ Z_{t}\in R^{N\times 2}$, $N$ is the number of key points which is determined by the facial alignment tool. 
The motion velocity and acceleration of key point at $i$ at timestep $t$ is formulated as:
\begin{equation}
v_{t}^{i} = Z_{t+1}^{i} - Z_{t}^{i},\ a_{t}^{i}= v_{t}^{i} - v_{t-1}^{i}
\end{equation}
where $v_{-1}^{i}=v_{0}^{i}=v_{T+1}^{i}=0$.

Intuitively, if the video suffers from significant motions jitters, the variations of motion will change dramatically, which means the statistical variance of the accelerations is much higher. The variance of the motion variations can be written as follows:
\begin{equation}
\label{equ:variance_i}
\sigma(a^{i}) = \frac{1}{\rm T-1}{\sum_{t}^{\rm T}(a_t^{i}-\bar{a}^{i})^2},
\end{equation}
where $\bar{a}^{i}$ is the mean of $a^{i}_{t}$ for $t=1,2,...,\rm T$. 
From our experimental results, the variance of accelerations is negatively correlated with the subjective score on motion stability (better motion stability, lower variance of acceleration). 
In this paper, we use the reciprocal of the variance of the acceleration as the motion stability index of the point. Our proposed objective metric for motion stability index is written as follows: 
\begin{equation}
\rm{MSI} (I_{1:T}) = \frac{1}{K}{\sum_{k=1}^{K}{(\sigma(a^k)+\epsilon)}^{-1}},
\end{equation}
where $\rm{K}$ is the number of selected key points, and $\epsilon=10^{-5}$. 

It is worth to mention that the scores of MSI in lips region differ from the scores in the jaws region which is because the different scale of motions in a face video. Hence, it is preferable to separately measure MSI in mouth region and jaw. 
To eliminate the interference of acceleration brought by different face sizes, we crop the video via unified strategy where a fixed bounding box is determined according to the ratio of the mouth width over the cropped image width. Specifically, we use the first 5 frames and the ratio of 0.25 to generate the cropping box of each video. For fair comparison, we rescale the face images into 256$\times$256 for each method.

\subsection{Loss Functions}
\label{subsec:loss_function}
Our neural renderer is optimized in supervised manner to generate a realistic image that are similar to the ground-truth image given with paired audio. We used two kinds of losses, namely reconstruction loss (unbalanced pixel-wise reconstruction loss and VGG perceptual loss~\cite{johnson2016perceptual}), and adversarial loss. The unbalanced reconstruction loss is employed to penalize the errors between the synthesized image sequence and the ground-truth image sequence (denoted as $\rm I_{1:T}$). 
To achieve this, we utilize the mouth masks (denoted as $\rm M_{1:T}$) to indicate different loss weights in the images where 2 for the weight in eroded region and 1 for the rest. A pre-trained VGG model is utilized to calculate the perceptual loss. The reconstrcution loss can be formulated as:
\begin{equation}
    \mathcal{L}_{rec} = \frac{1}{\rm T}\sum^{\rm T}_{t=1}(\|\rm \hat{I}_t\odot \rm M_t - \rm I_t\odot M_t\|_1+VGG(\rm\hat{I}_t, \rm I_t)),
\end{equation}
where the weight mask for mouth and non-mouth region are set to 2 and 1, respectively. 

The  adversarial losses for generator $G$ and discriminator $D$ in our work are implemented as the following formulation:
\begin{equation}
    \mathcal{L}_{adv}^{D} = -\mathbb{E}_{t\in [1,\rm T]}(\log D(\rm I_t)) - \mathbb{E}_{t\in[1,\rm T]}(\log (1 - D(\rm \hat{I}_t)),
\end{equation}
\begin{equation}
    \mathcal{L}_{adv}^{G} = \mathbb{E}_{t\in[1,\rm T]}(\log (1 - D(\rm \hat{I}_t)).
\end{equation}

In our experiment, $D$ shares the same structure as~\cite{thies2020neural}. The full objective of neural renderer is:
\begin{equation}
    \mathcal{L}_{total} = \lambda_1\mathcal{L}_{rec} + \lambda_2\mathcal{L}_{adv}^{G},
\end{equation}
where $\lambda_1$ and $\lambda_2$ are set to 20 and 1, respectively.

\section{Experiments and Results}
\label{sec:experiments}

\subsection{Experiment Settings}
\label{subsec:experiment_settings}

\noindent
\textbf{Dataset and Preprocessing.} We conduct experiments on three datasets, the GRID (an open-sourced standard dataset~\cite{cooke2006audio}, $\sim$50min for each person and 50\% for training), Testset 1 (collected from~\cite{guo2021adnerf}, $\sim$4min10s), and Testset 2 (collected from~\cite{suwajanakorn2017synthesizing}, $\sim$3min). 

\noindent
\textbf{Implementation Details.} 
We train our models in two stages, namely audio2expression and neural rendering. In the first stage, we extract the audio features and 3D facial expression parameters from the audio sequence and the video frames respectively. Since the mapping from audio to expression parameters are not our main contribution. we follow the settings of Transformer-S2A~\cite{chen2022transformers2a} and train an audio2expression model. In the second stage, treat a dozen of consecutive frames as a sequence, and train the neural renderer and adaptive smoothing module. We first optimize the neural renderer without adaptive smoothing module for 100 epochs, and then jointly train them together for another 20 epochs. Since the adaptive smoothing module is lightweight and contains only 11k learnable parameters, we optimize it with smaller learning rate (1e-6) than that of the neural renderer (1e-4). During the whole training procedure, we use augmented erosions for the input images and obtain the eroded images, and then concatenate them with the smoothed face shapes as the input of the neural renderer.

\noindent
\textbf{Comparison Settings.} 
\label{subsec:comparing_to_other_methods}
For the GRID dataset, we evaluate our method under self-reenactment setting (audios and videos are originally matched). We follow the setting of ~\cite{lahiri2021lipsync3d} and choose 10 subjects from GRID and use 50\% videos for training, 20\% for validating, and 30\% for testing. We compare our method to the results from previous competing methods (Vougioukas et. al.~\cite{vougioukas2020realistic}, ATVGnet~\cite{chen2019hierarchical}, Chen et. al.~\cite{chen2018lip} and LipSync3D~\cite{lahiri2021lipsync3d}). % We first train our StableFace in the training set, and then synthesize results under self-reenactment manner in which the audios are used to predict the facial expression parameters and the background face images are taken from the same talking video (similar to that in training). 
We adopt this evaluation setting since 1) it can easily calculate the objective evaluation metrics with the ground-truth target image (note that there is no ground-truth image in the common inference setting), and 2) it is convenient to compare with previous works since most of them adopt this evaluation setting.

For Testset 1 and 2, we extract speech audio from the demos of NVP~\cite{thies2020neural}, LipSync3D~\cite{lahiri2021lipsync3d}, LiveSP~\cite{lu2021live}, and AD-NeRF~\cite{guo2021adnerf} and animate the faces from Testset 1 and 2. Note that the audios and videos are not matched (without ground-truth videos), we cannot evaluate the results as we do in GRID. To this end, we conduct comparisons via both subjective experiments and objective metrics.

\begin{table}[!htbp]
  \footnotesize
  \caption{Comparisons with state-of-the-arts method on GRID~\cite{cooke2006audio}. }
  \centering
  \begin{tabular}{r|cccc}
    \toprule
    Method & SSIM$\uparrow$  & CPBD$\uparrow$ & WER$\downarrow$ & NLMD$\downarrow$ \\
    \hline
    Chen etal~\cite{chen2018lip} & 0.73  & 0.22& - & 0.018\\
    ATVGnet~\cite{chen2019hierarchical} & 0.83 & 0.17 & - & 0.011 \\
    Vougioukas et al.~\cite{vougioukas2020realistic} & 0.82  & 0.26 & 23\% & 0.010\\
    LipSync3D~\cite{lahiri2021lipsync3d} & 0.94  & 0.25 & {18}\% & \textbf{0.006}\\
    Ours & \textbf{0.97} & \textbf{0.35} & \textbf{12\%} & \textbf{0.006}\\
    \bottomrule
  \end{tabular}
  \label{tab:comparison_sotas}
\end{table}

\noindent
\textbf{Evaluating Metrics.} For comparisons on results with ground-truth images, we follow the setting of ~\cite{lahiri2021lipsync3d,vougioukas2020realistic} and adopt the metrics of: \textbf{SSIM}~\cite{wang2004image} to measure the quality of reconstructed image; \textbf{WER} (word error rate) to evaluate the accuracy of lip reading from the synthesized image using LipNet\cite{assael2016lipnet}\footnote{\url{https://github.com/Fengdalu/LipNet-PyTorch}}; \textbf{CPBD} (cumulative probability blur detection)~\cite{narvekar2009no} to evaluate the results sharpness; \textbf{NLMD} (normalized landmark distance that removes the influence of different input image resolution by dividing the landmark distance by image resolution, \emph{e.g.}, 256$\times$256) to measure the shape similarity of lips.

For comparison without ground-truth video setting in which the talking face videos are generated with audios from other source, we conduct both subjective and objective experiments. In subjective experiments, we choose MOS (mean opinion score) as our main rule. 
18 experienced users are invited to score the given videos at five grades according to perceived quality of motion stability\footnote{Are the motions around the lips and jaw stable across frames?}
, lip-sync quality\footnote{How does the motion of the lips match the speech in lip-audio accuracy?}
, and video realness\footnote{Does the video look real in your mind?}
. Different grade choices in MOS are: Very Good (5), Good (4), Average (3), Poor (2), Very Poor (1). A detailed annotation guideline and several real/generated videos are provided to ensure users can fully understand the difference between grades. Meanwhile, an additional labeling practice is conducted before the evaluation. For objective evaluation, we adopt \textbf{MSI} to measure the motion stability around lips and jaw regions respectively; \textbf{Sync-C} (SyncNet Confidence) to measure the lip-sync score and distance via SyncNet~\cite{chung2016out}.

\begin{figure}[!htp]
    \centering
    \includegraphics[width=1\linewidth]{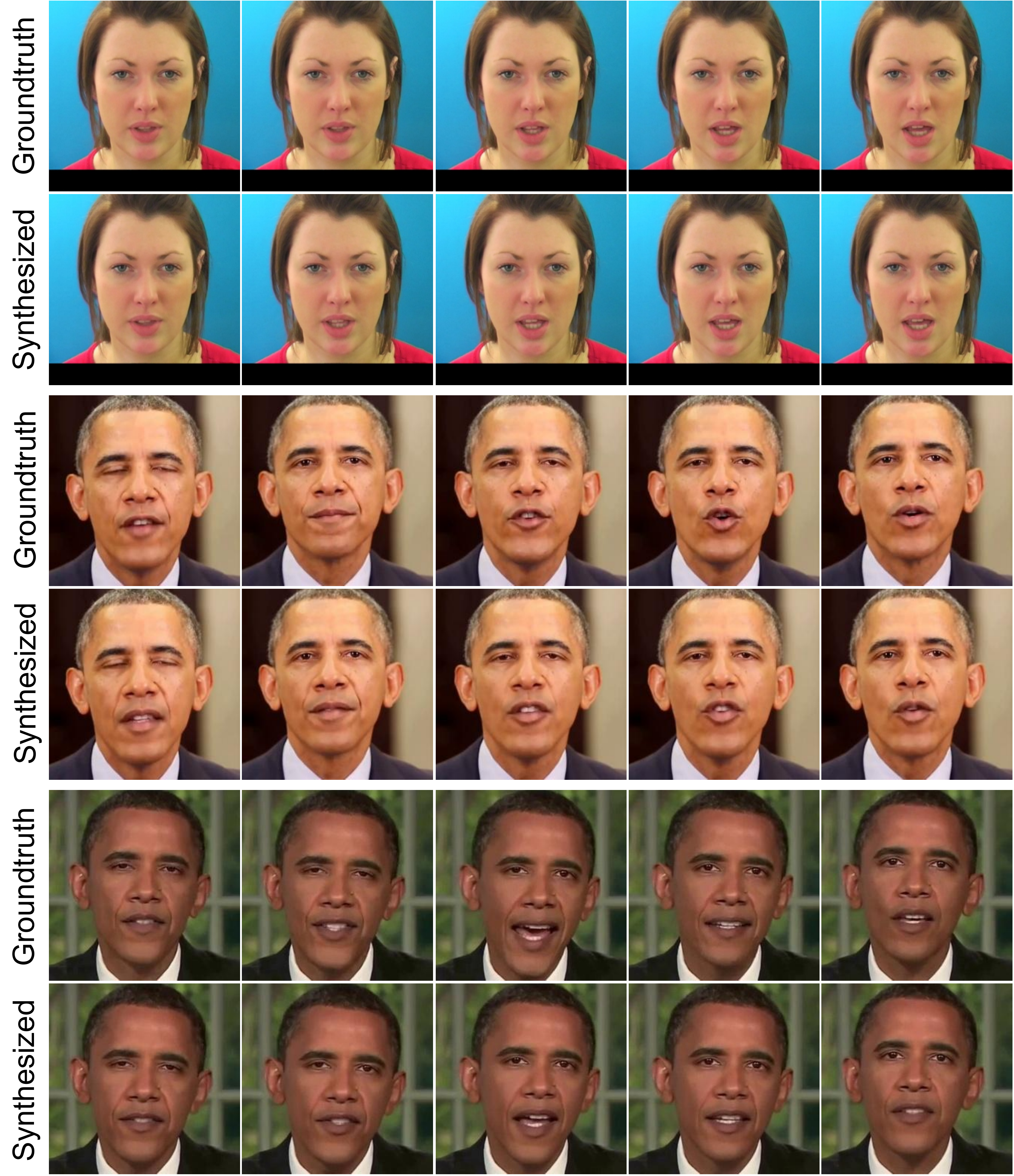}
    \caption{Given the speech as input, our method can synthesize high-fidelity results on GRID dataset and Obama videos.}
    \label{fig:representative_examples}
\end{figure}

\begin{table*}[!htbp]
  \small
  \caption{Quantitative comparisons with state-of-the-art methods. We measure the MOS score (higher means better) and the objective evaluation results on audio-driven generated videos. The best results except the real videos are marked in \textbf{bold}. }
  \centering
  \begin{tabular}{rccccccc}
    \toprule
    & \multicolumn{3}{c}{Subjective Evaluation} & \multicolumn{3}{c}{Objective Evaluation} \\
    \cmidrule(r){2-4}
    \cmidrule(r){5-7}
    {Method} & Motion Stability$\uparrow$ & Lip-sync Quality$\uparrow$ & Video Realness$\uparrow$ & MSI (Lip)$\uparrow$ & MSI (Jaw)$\uparrow$ & Sync-C$\uparrow$ \\% 
    \midrule
    Real videos & 4.31$\pm$0.25 & 4.70$\pm$0.13 & 4.59$\pm$0.13 & 0.450 & 0.690 & 6.92 \\ 
    \midrule
    NVP~\cite{thies2020neural} & 3.14$\pm$0.26 & 2.63$\pm$0.22 & {2.95}$\pm$0.23 & {0.399} & 0.588 & 4.55 \\ 
    LipSync3D~\cite{lahiri2021lipsync3d} & {3.05}$\pm$0.27 & 3.20$\pm$0.28 & {3.01}$\pm$0.22 & 0.384 & 0.598 & 5.40 \\ 
    LiveSP~\cite{lu2021live} & 2.47$\pm$0.26 & 2.56$\pm$0.27 & 2.61$\pm$0.23 & 0.414 & 0.756 & 5.02 \\
    AD-NeRF~\cite{guo2021adnerf} & 2.64$\pm$0.30 & 3.15$\pm$0.27 & 2.65$\pm$0.27 & 0.348 & 0.490 & 5.14 \\ 
    Ours & \textbf{3.72$\pm$0.22} & \textbf{3.60$\pm$0.21} & \textbf{3.68$\pm$0.20} & \textbf{0.447} & \textbf{0.752} & \textbf{5.62} \\ 
    \bottomrule
  \end{tabular}
  \label{tab:user_study_for_methods}
\end{table*}

\begin{table}[!htbp]
  \footnotesize
  \centering
  \caption{Pearson correlation coefficients for the subjective measures and objective metrics. }
  \begin{tabular}{c|cccccc}
    \toprule
     {\diagbox[]{Subjective}{Objective}} & $\sigma(v)$ & $1/\sigma(v)$ & $\sigma(a)$ & {MSI} & \\
    \hline
    \multirow{2}{*}{Motion Stability Score}  & -0.157 & {0.333} & -0.191 & \textbf{0.424} & (Lip) \\
       & -0.153 & {0.378} & -0.163 & \textbf{0.438} & (Jaw) \\
    \bottomrule
  \end{tabular}
  \label{tab:correlation_MSI_MOS_score}
\end{table}

\begin{figure*}[!htbp]
    \centering
    \includegraphics[width=0.9\linewidth]{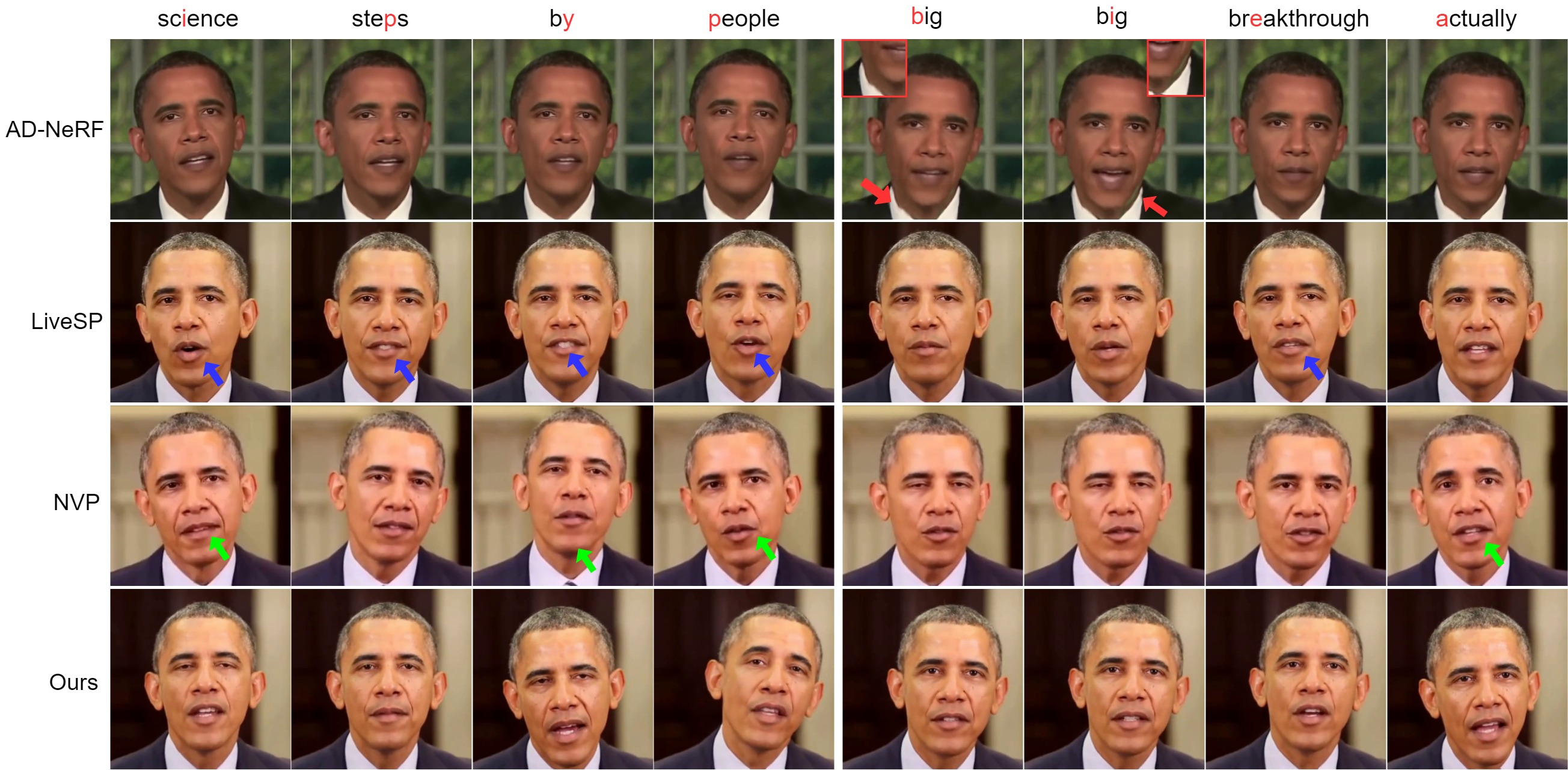}
    \caption{Qualitative comparison on Testset 2. The images of other methods are cropped from their demos. Our method not only preserves the high quality of face images, but also generates videos with better lip-sync quality. On the contrary, AD-NeRF~\cite{guo2021adnerf} suffers from generating images with cracks between the head and torso (\textcolor{red}{red arrow}). LiveSP~\cite{lu2021live} synthesizes videos where lips are less sync with the audios (\textcolor{blue}{blue arrow}) than our method, while NVP~\cite{thies2020neural} generates videos with nearly muted lip movements and blurry details(\textcolor{green}{green arrow}). }
    \label{fig:qualitative_comparison}
\end{figure*}

\subsection{Results}
\label{subsec:results}
The quantitative results of \underline{comparison with ground-truth} setting are presented in Table~\ref{tab:comparison_sotas}. It can be concluded that our method outperforms state-of-the-art methods in all objective metrics, \emph{i.e.}, SSIM, CPBD, and WER, on the GRID dataset. With NLMD which measures the normalized landmark distance between synthesized images and the ground-truth, we eleminate the errors that may be brought by image resolution, and present the comparison results to fairly demonstrate the better performance of our method on self-reenactment. 
In terms of NLMD which, we achieve comparable performance to LipSync3D. Note that LipSync3D generates face images in resolution of 128$\times$128, while our method work on images of 256$\times$256. Our synthesized videos have better quality of sharpness and less word error rate. 
In Fig.~\ref{fig:representative_examples}, we visualize some key-frame results of our methods on GRID and Obama's video. Our method can achieve plausible results with shapr details and high image quality under different illumination conditions, skin color, and \emph{etc}.

We proceed to evaluate the efficacy of our MSI metric. In Table~\ref{tab:correlation_MSI_MOS_score}, we present the Pearson correlation coefficients between our MSI and the subjective scores of motion stability, where the MSI results and subjective scores of motion stability on 99 videos. 
Instead of measuring motion stability via the variance of motions or variance of accelerations across frames, we take the reciprocal value of the motion variations to achieve higher correlations between objective metric and subjective scores on motion stability. 
It can be found that the correlation coefficients between MSI metric and motion stability in videos reach to 0.424 and 0.438 for lip and jaw respectively, which are higher than the statistical measures.

For \underline{comparison without groudtruth videos}, we present our quantitative results in Table~\ref{tab:user_study_for_methods}. In the left of the table, we show the statistical MOS in subjective study. Different from previous works that only evaluate the frame visual quality and lip-sync quality, we add a new evaluating item and show that our synthesized videos looks much better than previous other works. In the right of the table, we evaluate those videos with objective metrics. 
Considering that different scale of motions around the lips and jaw in talking face video, we separately measure MSI for the lips and jaw for each video. We first compare the results of motion stability, our method achieves higher MOS of 3.72 compared to the second best results. From the results of MSI (Lip) and MSI (Jaw), our method consistently outperforms the competing methods, demonstrating the effectiveness of our solutions on solving the motion jittering problem in talking face generation task. According to the results of motion stability and video realness, it can be concluded that the method that generates videos with better motion stability tends to attain better results in video realness.

\begin{table*}[!htp]
\caption{Ablation studies on each component in our proposed method. }
\footnotesize
\begin{tabular}{rcccccccccccc}
\toprule
 & \multicolumn{5}{c}{Testset 1} & \multicolumn{5}{c}{Testset 2}\\
\cmidrule(r){2-6}
\cmidrule(r){7-11}
{Method} & SSIM$\uparrow$ & Sync-C$\uparrow$ & NLMD$\downarrow$  & MSI-Lip$\uparrow$ & MSI-Jaw$\uparrow$ & SSIM$\uparrow$ & Sync-C$\uparrow$ & NLMD$\downarrow$ & MSI-Lip$\uparrow$ & MSI-Jaw$\uparrow$ \\
\midrule
Full Model & \textbf{0.939} & \textbf{5.77} & \textbf{0.0108} & 0.504 & \textbf{0.997} & \textbf{0.964} & \textbf{5.31}  & \textbf{0.0098} & \textbf{0.565} & {1.005} \\

w/o Audio Fusion Module & 0.930 & 5.59 & 0.0126 & \textbf{0.510} & 0.994 & 0.956 & 5.05 & 0.0110 & 0.560 & \textbf{1.010} \\

w/o Adaptive Smoothing  & {0.935} & {5.80}  & {0.0117} & 0.470 & 0.954 & {0.960} & 5.29  & 0.0105 & 0.519 & 0.940\\

w/o Augmented Erosion & 0.932  & 5.74  & 0.0121 & 0.495 & 0.985 & 0.955& 5.26  & 0.0101 & 0.535 & 0.993 \\

w/o Dependency Modeling & 0.929 & 5.51 & 0.0130  & 0.439 & 0.926 & 0.948  & 4.57  & 0.0119 & 0.526 & 0.922\\

Baseline Model & 0.928 & 5.37  & 0.0130 & 0.378 & 0.776 & 0.950 & 4.69  & 0.0123 & 0.503 & 0.857\\
\bottomrule
\end{tabular}
\label{tab:ablation_objective}
\end{table*}

To qualitatively compare the generated results of each method, we extract key frames from the videos of each method and show the results in Fig.~\ref{fig:qualitative_comparison}. Among the images synthesized by AD-NeRF~\cite{guo2021adnerf}, some of them suffer from clear \textbf{cracks} between head and torso. The reason is that AD-NeRF synthesizes the head and torso separately, without considering spatial consistency within each frame. Meanwhile, from their video results suffer from severe body jitters. NVP~\cite{thies2020neural} generates videos with nearly muted lip movements while LiveSP~\cite{lu2021live} fails to synthesizes accurate lips in sync with the audios. In contrast, our method predicts more accurate facial expressions, and generates videos with higher quality and more accurate lip movements. (We strongly suggest reviewers to watch the enclosed video for better visual comparison experiences on motion jitters.)

\begin{figure}[!htbp]
    \centering
    \includegraphics[width=1\linewidth]{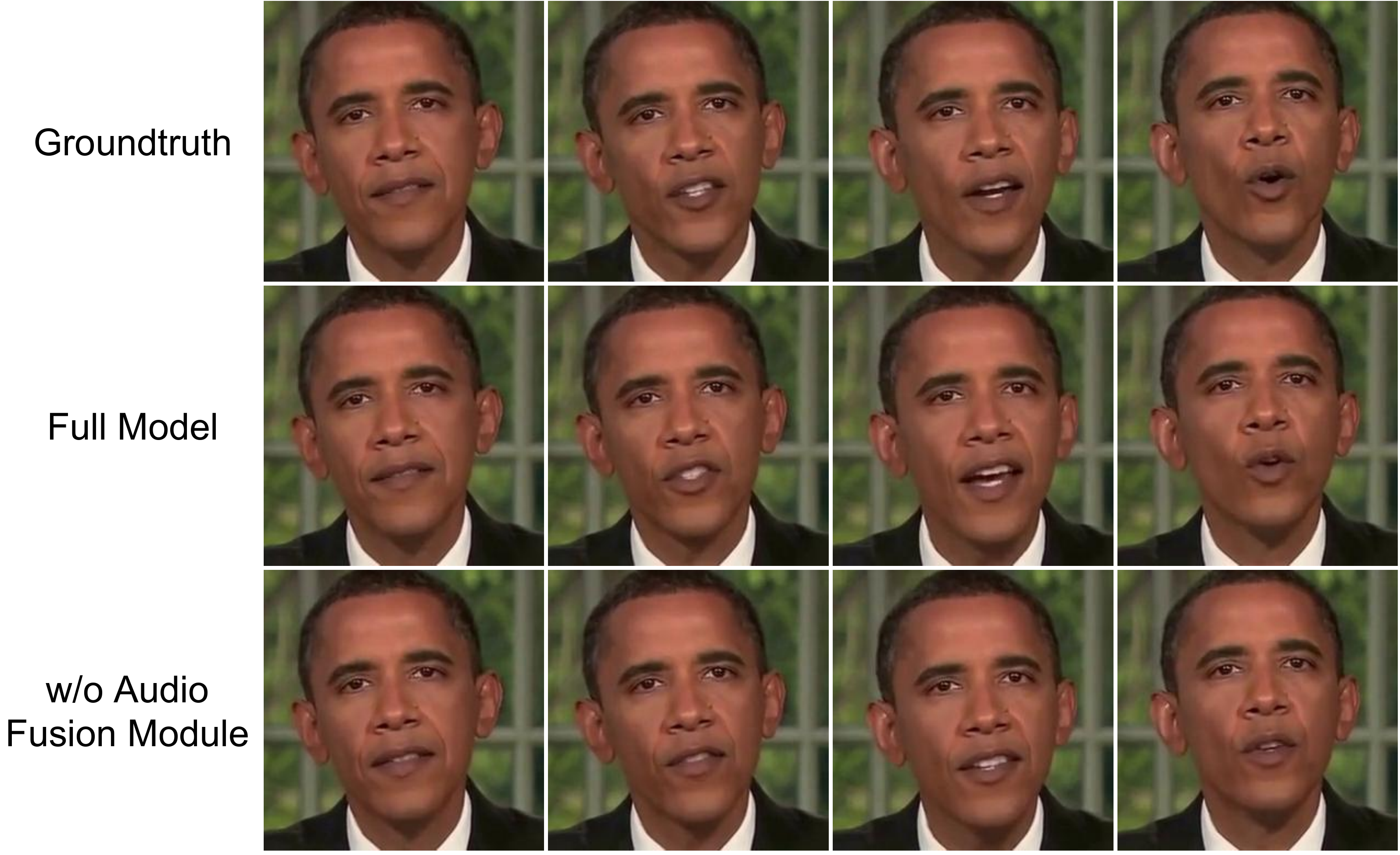}
    \caption{Ablation study on audio fusion module. Please zoom in for better observation. }
    \label{fig:ablation_wo_audio}
\end{figure}

\subsection{Ablation Studies}
\label{subsec:ablation_studies}
To rationalize each designing choice of our method, we conduct extensive ablation experiments by removing each component or all of our three components. From the comparison results of Sync-C in Table~\ref{tab:ablation_objective}, we find that our dependency module helps to improve the lip-sync quality on both datasets. Meanwhile, in evaluating image similarity with SSIM and shape similarity with NLMD, our full model achieves clear improvements over the method without augmented erosion, dependency module, or adaptive smoothing module. It can be seen that the SSIM results get worse after removing each of our proposed components, which verifies the effectiveness of each design.

\begin{figure}[!htbp]
    \centering
    \includegraphics[width=1\linewidth]{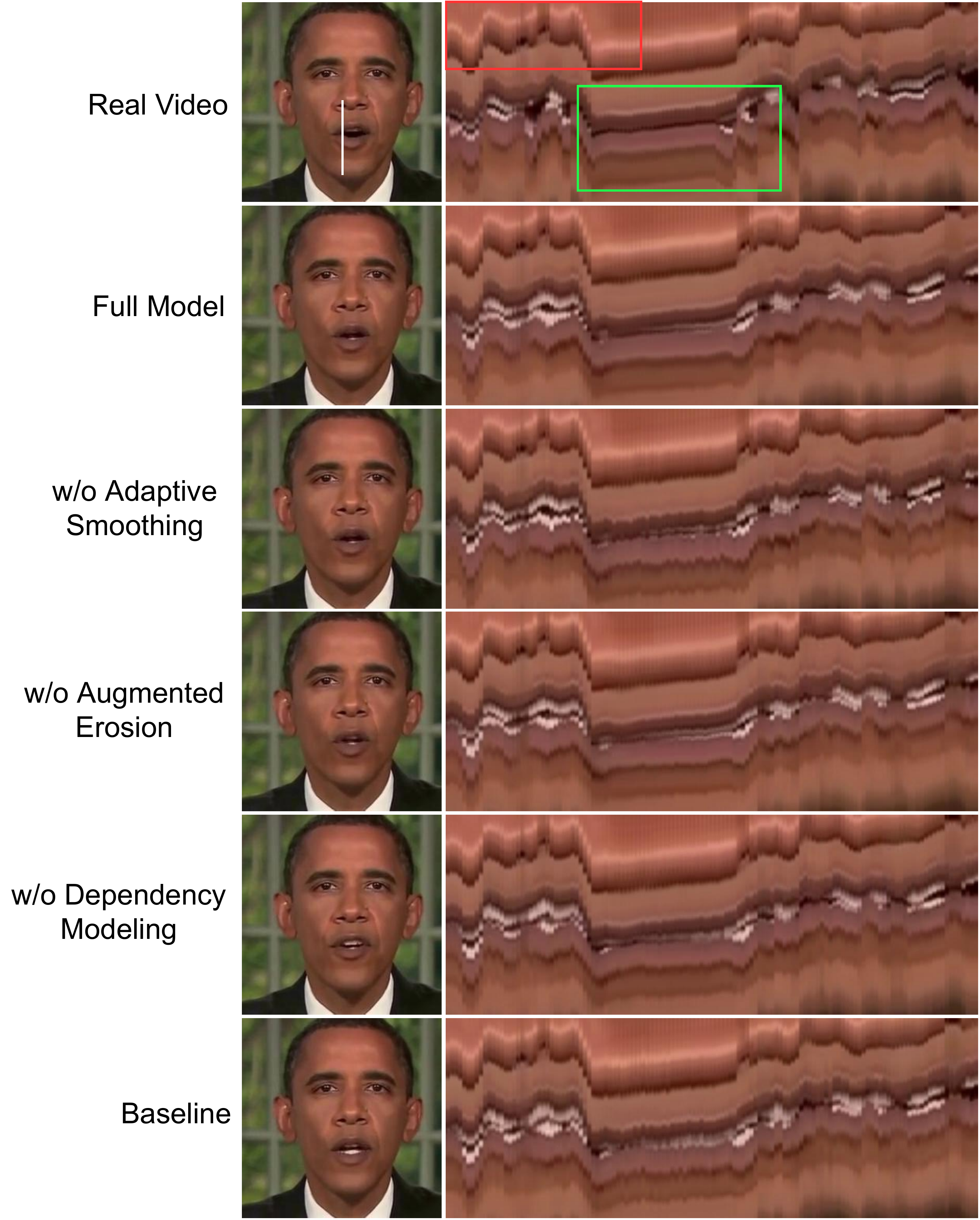}
    \caption{The visualization of vertical slice along the video frames. From top to bottom, we slice each frames at a fixed horizontal location to visualize the jittering patterns. }
    \label{fig:ablation_visualization}
\end{figure}

We compare our full model with the one without audio fusion. Fig.~\ref{fig:ablation_wo_audio} gives the comparison results, from which we conclude that audio features have the better capability in helping the neural renderer synthesize more accurate lips. In Fig.~\ref{fig:ablation_visualization}, we concatenate the vertical slice in each frame along the time for each video. The position and size of the vertical slice are shown in the first row and kept the same for each video. We have two observations from the figure. First, without each component, we can observe some jitter patterns from the vertical slice figure. Compared to the baseline, the vertical slip figure of the full model is much more smooth and close to the real video. Meanwhile, the jaggy patterns can also be observed compared to other methods without each component. Second, it may be concluded from the figure that the jitters are more likely to occur in the lip regions (green box) than the nose region (red box) because the lip motions are much more than the nose which is almost static relative to the head. 

\begin{figure}[!htbp]
  \centering
  \begin{subfigure}[Comparison with w/o adaptive smoothing.]{
    \includegraphics[width=0.98\linewidth]{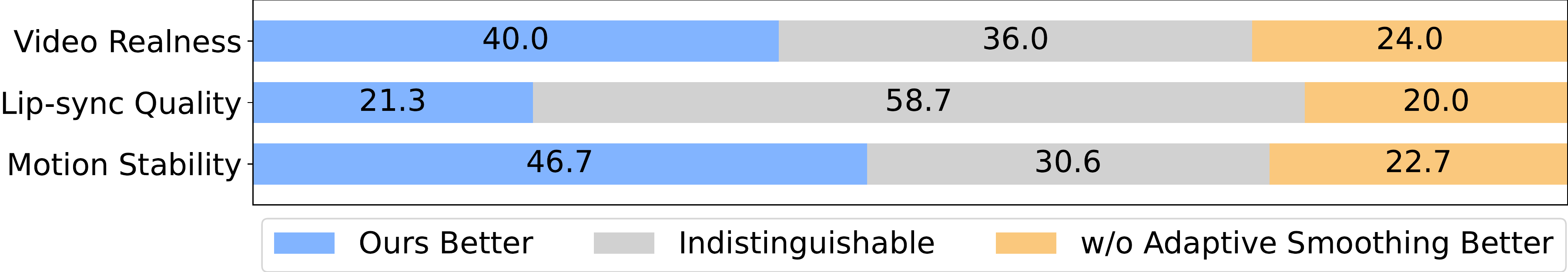}
    }
  \end{subfigure}\\
  \vspace{-0.24cm}
  \centering
  \begin{subfigure}[Comparison with w/o augmented erosion.]{
    \includegraphics[width=0.98\linewidth]{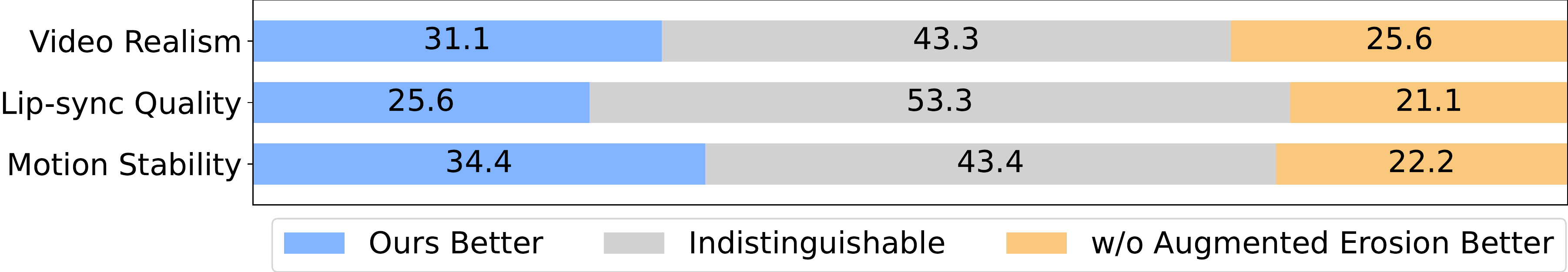}
    }
  \end{subfigure}\\
  \vspace{-0.24cm}
  \centering
  \begin{subfigure}[Comparison with w/o dependency modeling.]{
    \includegraphics[width=0.98\linewidth]{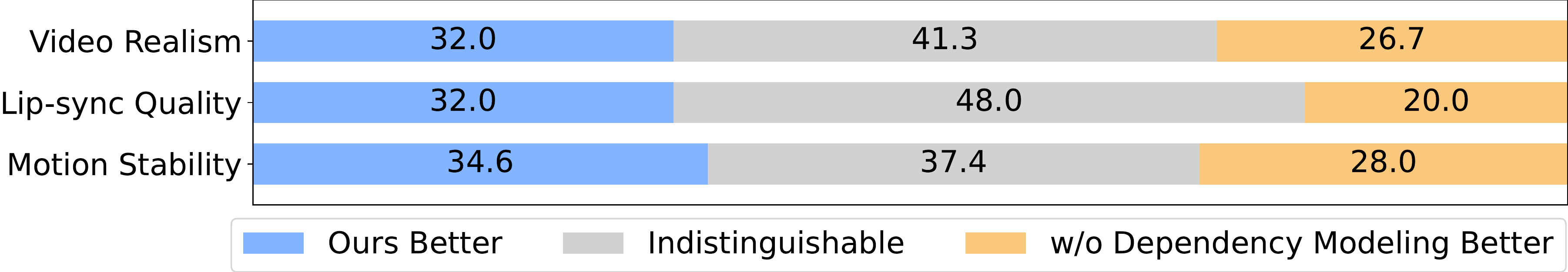}
  }
  \end{subfigure}\\
  % \vspace{-0.5cm}
  \caption{Ablation studies on the three components in our proposed method with subjective evaluation (user preference test).}
  \label{fig:ablation_study}
\end{figure}

\begin{figure}[!htbp]
  \centering
  \begin{subfigure}[Comparison with fixed smoothing weight. ]{
    \includegraphics[width=0.98\linewidth]{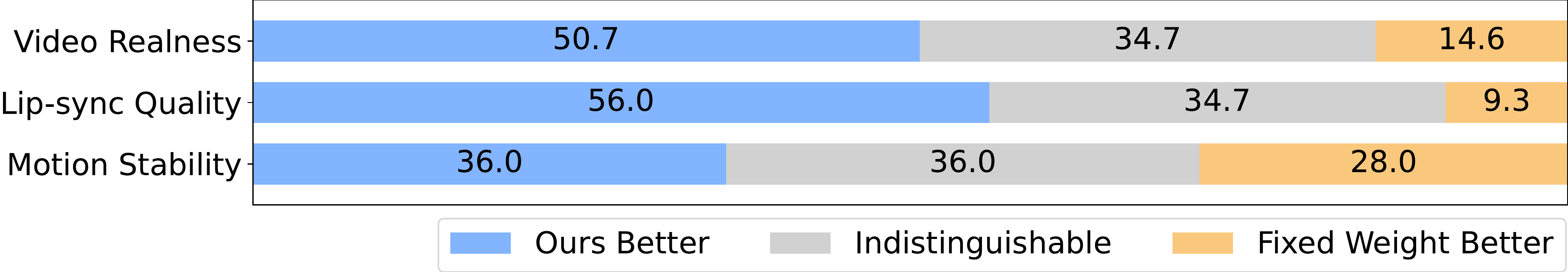}
    }
    \label{fig:ablation_subfig_a}
  \end{subfigure}\\
  \vspace{-0.24cm}
  \centering
  \begin{subfigure}[Comparison with global but learnable smoothing weight.]{
    \includegraphics[width=0.98\linewidth]{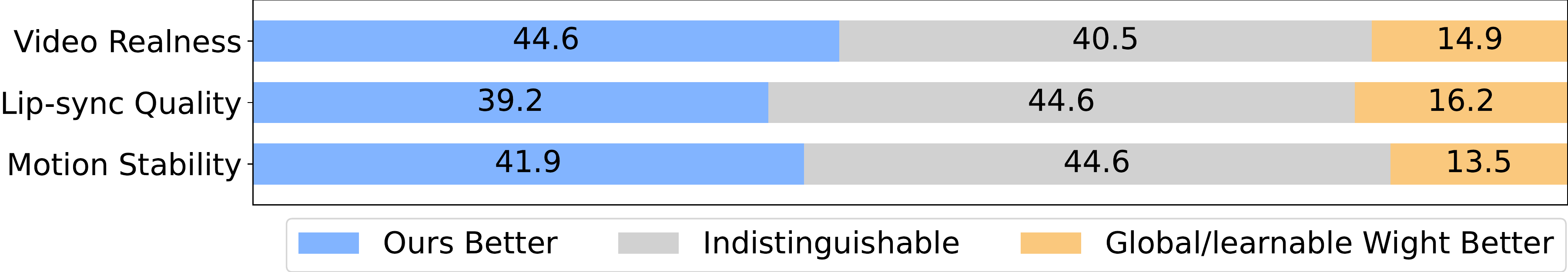}
    }
    \label{fig:ablation_subfig_b}
  \end{subfigure}\\
  \caption{Comparison on different choices for adaptive smoothing: adaptive smoothing weight, fixed smoothing weight, and global/learnable weight (user preference test). Our adaptive smoothing design outperforms other choices by a large margin. }
  \label{fig:adaptive_smoothing_analyses}
\end{figure}

In Fig.~\ref{fig:ablation_study}, we visualize the user preference results in terms of motion stability, lip-sync quality, and realness. From each sub-figure, it can be observed that the preference percentage motion stability reduces when we remove the adaptive smoothing module, dependency modeling or augmented erosion. 
Fig.~\ref{fig:adaptive_smoothing_analyses} shows the user preference studies over different smoothing strategy. From the results, we can tell that our adaptive smoothing choice clearly attains more votes than that of fixed weight or global but learnable weight, especially on lip-sync quality and video realness.

%-------------------------------------------------------------------------
\section{Discussion}
\label{sec:discussions}
Our technology of photo-realistic talking face generation aims for goodness, value, and productivity in multimedia applications. 
For instance, video re-dubbing has a potentially beneficial promotion in online education in different countries and regions, no matter what the original spoken language is. Our technology is also capable of synthesizing news broadcasting videos under supervision. To benefit the community, our developed technique could also be used to synthesize fake videos as negative samples of deepfake dataset. 

Despite its potential benefits, ethical risk is also raised since these techniques can be potentially misused and do harm. We strongly suggest the usage of our technology should be under the supervision and any videos generated by our method should be labeled with clearly visible watermarks, and the request for code would be carefully assessed. We believe that with the joint efforts of the research community and the industry, we would find a promising way to foster the technology while maintaining Pandora's box close.

\section{Conclusion}
In this paper, we develop a motion-stable talking face generation system. 
We systematically analyze the motion jittering problem and then propose three key solutions to improve the motion stability of synthesized videos. 
Moreover, we propose an effective metric (MSI). To our best knowledge, MSI is the first objective metric that evaluates the motion stability in talking face videos. Extensive experiments, including both objective and subjective evaluations, are conducted to demonstrate the effectiveness of our proposed solutions in improving motion stability of the talking face videos. 
For future work, we will extend to more application scenarios, such as emotional talking face generation, which is more challenging for motion stability but also makes talking face video more expressive.

% -------------------------------------------------------------------------------------
% user defined bib files
\bibliographystyle{IEEEtran}
% % argument is your BibTeX string definitions and bibliography database(s)
\bibliography{acmart}

\begin{thebibliography}{10}
\providecommand{\url}[1]{#1}
\csname url@samestyle\endcsname
\providecommand{\newblock}{\relax}
\providecommand{\bibinfo}[2]{#2}
\providecommand{\BIBentrySTDinterwordspacing}{\spaceskip=0pt\relax}
\providecommand{\BIBentryALTinterwordstretchfactor}{4}
\providecommand{\BIBentryALTinterwordspacing}{\spaceskip=\fontdimen2\font plus
\BIBentryALTinterwordstretchfactor\fontdimen3\font minus
  \fontdimen4\font\relax}
\providecommand{\BIBforeignlanguage}[2]{{%
\expandafter\ifx\csname l@#1\endcsname\relax
\typeout{** WARNING: IEEEtran.bst: No hyphenation pattern has been}%
\typeout{** loaded for the language `#1'. Using the pattern for}%
\typeout{** the default language instead.}%
\else
\language=\csname l@#1\endcsname
\fi
#2}}
\providecommand{\BIBdecl}{\relax}
\BIBdecl

\bibitem{chen2019hierarchical}
L.~Chen, R.~K. Maddox, Z.~Duan, and C.~Xu, ``Hierarchical cross-modal talking
  face generation with dynamic pixel-wise loss,'' in \emph{Proc. CVPR}, 2019,
  pp. 7832--7841.

\bibitem{zhou2020makelttalk}
Y.~Zhou, X.~Han, E.~Shechtman, J.~Echevarria, E.~Kalogerakis, and D.~Li,
  ``Makelttalk: speaker-aware talking-head animation,'' \emph{ACM TOG},
  vol.~39, no.~6, pp. 1--15, 2020.

\bibitem{prajwal2020lip}
K.~Prajwal, R.~Mukhopadhyay, V.~P. Namboodiri, and C.~Jawahar, ``A lip sync
  expert is all you need for speech to lip generation in the wild,'' in
  \emph{Proc. ACM MM}, 2020, pp. 484--492.

\bibitem{vougioukas2020realistic}
K.~Vougioukas, S.~Petridis, and M.~Pantic, ``Realistic speech-driven facial
  animation with gans,'' \emph{IJCV}, vol. 128, no.~5, pp. 1398--1413, 2020.

\bibitem{zhou2021pose}
H.~Zhou, Y.~Sun, W.~Wu, C.~C. Loy, X.~Wang, and Z.~Liu, ``Pose-controllable
  talking face generation by implicitly modularized audio-visual
  representation,'' in \emph{Proc. CVPR}, 2021, pp. 4176--4186.

\bibitem{wu2021imitating}
H.~Wu, J.~Jia, H.~Wang, Y.~Dou, C.~Duan, and Q.~Deng, ``Imitating arbitrary
  talking style for realistic audio-driven talking face synthesis,'' in
  \emph{Proc. ACM MM}, 2021, pp. 1478--1486.

\bibitem{chen2020duallip}
W.~Chen, X.~Tan, Y.~Xia, T.~Qin, Y.~Wang, and T.-Y. Liu, ``Duallip: A system
  for joint lip reading and generation,'' in \emph{Proc. ACM MM}, 2020, pp.
  1985--1993.

\bibitem{chen2020talking}
L.~Chen, G.~Cui, C.~Liu, Z.~Li, Z.~Kou, Y.~Xu, and C.~Xu, ``Talking-head
  generation with rhythmic head motion,'' in \emph{Proc. ECCV}.\hskip 1em plus
  0.5em minus 0.4em\relax Springer, 2020, pp. 35--51.

\bibitem{yi2020audio}
R.~Yi, Z.~Ye, J.~Zhang, H.~Bao, and Y.-J. Liu, ``Audio-driven talking face
  video generation with learning-based personalized head pose,'' \emph{arXiv
  preprint arXiv:2002.10137}, 2020.

\bibitem{ji2021audio}
X.~Ji, H.~Zhou, K.~Wang, W.~Wu, C.~C. Loy, X.~Cao, and F.~Xu, ``Audio-driven
  emotional video portraits,'' in \emph{Proc. CVPR}, 2021, pp.
  14\,080--14\,089.

\bibitem{fried2019text}
O.~Fried, A.~Tewari, M.~Zollh{\"o}fer, A.~Finkelstein, E.~Shechtman, D.~B.
  Goldman, K.~Genova, Z.~Jin, C.~Theobalt, and M.~Agrawala, ``Text-based
  editing of talking-head video,'' \emph{ACM TOG}, vol.~38, no.~4, pp. 1--14,
  2019.

\bibitem{yao2021iterative}
X.~Yao, O.~Fried, K.~Fatahalian, and M.~Agrawala, ``Iterative text-based
  editing of talking-heads using neural retargeting,'' \emph{ACM TOG}, vol.~40,
  no.~3, pp. 1--14, 2021.

\bibitem{seshadrinathan2009motion}
K.~Seshadrinathan and A.~C. Bovik, ``Motion-based perceptual quality assessment
  of video,'' in \emph{Human Vision and Electronic Imaging XIV}, vol.
  7240.\hskip 1em plus 0.5em minus 0.4em\relax International Society for Optics
  and Photonics, 2009, p. 72400X.

\bibitem{thies2020neural}
J.~Thies, M.~Elgharib, A.~Tewari, C.~Theobalt, and M.~Nie{\ss}ner, ``Neural
  voice puppetry: Audio-driven facial reenactment,'' in \emph{Proc.
  ECCV}.\hskip 1em plus 0.5em minus 0.4em\relax Springer, 2020, pp. 716--731.

\bibitem{guo2021adnerf}
Y.~Guo, K.~Chen, S.~Liang, Y.-J. Liu, H.~Bao, and J.~Zhang, ``Ad-nerf: Audio
  driven neural radiance fields for talking head synthesis,'' in \emph{Proc.
  ICCV}, October 2021, pp. 5784--5794.

\bibitem{lu2021live}
Y.~Lu, J.~Chai, and X.~Cao, ``{Live Speech Portraits}: Real-time photorealistic
  talking-head animation,'' \emph{ACM Transactions on Graphics}, vol.~40,
  no.~6, December 2021.

\bibitem{lahiri2021lipsync3d}
A.~Lahiri, V.~Kwatra, C.~Frueh, J.~Lewis, and C.~Bregler, ``Lipsync3d:
  Data-efficient learning of personalized 3d talking faces from video using
  pose and lighting normalization,'' in \emph{Proc. CVPR}, 2021, pp.
  2755--2764.

\bibitem{wang2004image}
Z.~Wang, A.~C. Bovik, H.~R. Sheikh, and E.~P. Simoncelli, ``Image quality
  assessment: from error visibility to structural similarity,'' \emph{TIP},
  vol.~13, no.~4, pp. 600--612, 2004.

\bibitem{chen2018lip}
L.~Chen, Z.~Li, R.~K. Maddox, Z.~Duan, and C.~Xu, ``Lip movements generation at
  a glance,'' in \emph{Proc. ECCV}, 2018, pp. 520--535.

\bibitem{song2021tacr}
L.~Song, B.~Liu, G.~Yin, X.~Dong, Y.~Zhang, and J.-X. Bai, ``Tacr-net: Editing
  on deep video and voice portraits,'' in \emph{Proc. ACM MM}, 2021, pp.
  478--486.

\bibitem{wen2020photorealistic}
X.~Wen, M.~Wang, C.~Richardt, Z.-Y. Chen, and S.-M. Hu, ``Photorealistic
  audio-driven video portraits,'' \emph{TVCG}, vol.~26, no.~12, pp. 3457--3466,
  2020.

\bibitem{zhang2021facial}
C.~Zhang, Y.~Zhao, Y.~Huang, M.~Zeng, S.~Ni, M.~Budagavi, and X.~Guo, ``Facial:
  Synthesizing dynamic talking face with implicit attribute learning,'' in
  \emph{Proc. ICCV}, 2021, pp. 3867--3876.

\bibitem{feng2021deca}
Y.~Feng, H.~Feng, M.~J. Black, and T.~Bolkart, ``Learning an animatable
  detailed 3d face model from in-the-wild images,'' \emph{ACM TOG}, vol.~40,
  no.~4, pp. 1--13, 2021.

\bibitem{thies2016face2face}
J.~Thies, M.~Zollhofer, M.~Stamminger, C.~Theobalt, and M.~Nie{\ss}ner,
  ``Face2face: Real-time face capture and reenactment of rgb videos,'' in
  \emph{Proc. CVPR}, 2016, pp. 2387--2395.

\bibitem{ling2020toward}
J.~Ling, H.~Xue, L.~Song, S.~Yang, R.~Xie, and X.~Gu, ``Toward fine-grained
  facial expression manipulation,'' in \emph{Proc. ECCV}.\hskip 1em plus 0.5em
  minus 0.4em\relax Springer, 2020, pp. 37--53.

\bibitem{wiles2018x2face}
O.~Wiles, A.~Koepke, and A.~Zisserman, ``X2face: A network for controlling face
  generation using images, audio, and pose codes,'' in \emph{Proc. ECCV}, 2018,
  pp. 670--686.

\bibitem{kim2018deep}
H.~Kim, P.~Garrido, A.~Tewari, W.~Xu, J.~Thies, M.~Niessner, P.~P{\'e}rez,
  C.~Richardt, M.~Zollh{\"o}fer, and C.~Theobalt, ``Deep video portraits,''
  \emph{ACM TOG}, vol.~37, no.~4, pp. 1--14, 2018.

\bibitem{nagano2018pagan}
K.~Nagano, J.~Seo, J.~Xing, L.~Wei, Z.~Li, S.~Saito, A.~Agarwal, J.~Fursund,
  and H.~Li, ``pagan: real-time avatars using dynamic textures,'' \emph{ACM
  TOG}, vol.~37, no.~6, pp. 1--12, 2018.

\bibitem{xue2020realistic}
H.~Xue, J.~Ling, L.~Song, R.~Xie, and W.~Zhang, ``Realistic talking face
  synthesis with geometry-aware feature transformation,'' in \emph{Proc.
  ICIP}.\hskip 1em plus 0.5em minus 0.4em\relax IEEE, 2020, pp. 1581--1585.

\bibitem{tang2022generative}
A.~Tang, Y.~Huang, J.~Ling, Z.~Zhang, Y.~Zhang, R.~Xie, and L.~Song,
  ``Generative compression for face video: A hybrid scheme,'' in \emph{Proc.
  ICME}, 2022.

\bibitem{chung2017you}
J.~S. Chung, A.~Jamaludin, and A.~Zisserman, ``You said that?'' in \emph{Proc.
  BMVC}, 2017.

\bibitem{zhou2019talking}
H.~Zhou, Y.~Liu, Z.~Liu, P.~Luo, and X.~Wang, ``Talking face generation by
  adversarially disentangled audio-visual representation,'' in \emph{Proc.
  AAAI}, vol.~33, no.~01, 2019, pp. 9299--9306.

\bibitem{suwajanakorn2017synthesizing}
S.~Suwajanakorn, S.~M. Seitz, and I.~Kemelmacher-Shlizerman, ``Synthesizing
  obama: learning lip sync from audio,'' \emph{ACM TOG}, vol.~36, no.~4, pp.
  1--13, 2017.

\bibitem{siarohin2019first}
A.~Siarohin, S.~Lathuili{\`e}re, S.~Tulyakov, E.~Ricci, and N.~Sebe, ``First
  order motion model for image animation,'' in \emph{Proc. NeurIPS}, 2019, pp.
  7137--7147.

\bibitem{wang2019few}
T.-C. Wang, M.-Y. Liu, A.~Tao, G.~Liu, B.~Catanzaro, and J.~Kautz, ``Few-shot
  video-to-video synthesis,'' in \emph{Proc. NeurIPS}, 2019, pp. 5013--5024.

\bibitem{zakharov2019few}
E.~Zakharov, A.~Shysheya, E.~Burkov, and V.~Lempitsky, ``Few-shot adversarial
  learning of realistic neural talking head models,'' in \emph{Proc. ICCV},
  2019, pp. 9459--9468.

\bibitem{zakharov2020fast}
E.~Zakharov, A.~Ivakhnenko, A.~Shysheya, and V.~Lempitsky, ``Fast bi-layer
  neural synthesis of one-shot realistic head avatars,'' in \emph{Proc.
  ECCV}.\hskip 1em plus 0.5em minus 0.4em\relax Springer, 2020, pp. 524--540.

\bibitem{blanz1999morphable}
V.~Blanz and T.~Vetter, ``A morphable model for the synthesis of 3d faces,'' in
  \emph{Proceedings of the 26th annual conference on Computer graphics and
  interactive techniques}, 1999, pp. 187--194.

\bibitem{doukas2021headgan}
M.~C. Doukas, S.~Zafeiriou, and V.~Sharmanska, ``Headgan: One-shot neural head
  synthesis and editing,'' in \emph{Proc. ICCV}, 2021, pp. 14\,398--14\,407.

\bibitem{vaswani2017attention}
A.~Vaswani, N.~Shazeer, N.~Parmar, J.~Uszkoreit, L.~Jones, A.~N. Gomez,
  {\L}.~Kaiser, and I.~Polosukhin, ``Attention is all you need,'' in
  \emph{Proc. NeurIPS}, 2017, pp. 5998--6008.

\bibitem{devlin2018bert}
J.~Devlin, M.-W. Chang, K.~Lee, and K.~Toutanova, ``Bert: Pre-training of deep
  bidirectional transformers for language understanding,'' \emph{arXiv preprint
  arXiv:1810.04805}, 2018.

\bibitem{ren2019fastspeech}
Y.~Ren, Y.~Ruan, X.~Tan, T.~Qin, S.~Zhao, Z.~Zhao, and T.-Y. Liu, ``Fastspeech:
  Fast, robust and controllable text to speech,'' 2019, pp. 3171--3180.

\bibitem{fan2022faceformer}
Y.~Fan, Z.~Lin, J.~Saito, W.~Wang, and T.~Komura, ``Faceformer: Speech-driven
  3d facial animation with transformers,'' in \emph{Proc. CVPR}, 2022, pp.
  18\,770--18\,780.

\bibitem{dosovitskiy2020image}
A.~Dosovitskiy, L.~Beyer, A.~Kolesnikov, D.~Weissenborn, X.~Zhai,
  T.~Unterthiner, M.~Dehghani, M.~Minderer, G.~Heigold, S.~Gelly \emph{et~al.},
  ``An image is worth 16x16 words: Transformers for image recognition at
  scale,'' in \emph{ICLR}, 2021.

\bibitem{chen2022transformers2a}
L.~Chen, Z.~Wu, J.~Ling, R.~Li, X.~Tan, and S.~Zhao, ``Transformer-s2a: Robust
  and efficient speech-to-animation,'' in \emph{Proc. ICASSP}.\hskip 1em plus
  0.5em minus 0.4em\relax IEEE, 2022, pp. 7247--7251.

\bibitem{liu2021fuseformer}
R.~Liu, H.~Deng, Y.~Huang, X.~Shi, L.~Lu, W.~Sun, X.~Wang, J.~Dai, and H.~Li,
  ``Fuseformer: Fusing fine-grained information in transformers for video
  inpainting,'' in \emph{Proc. ICCV}, 2021, pp. 14\,040--14\,049.

\bibitem{ren2022dlformer}
J.~Ren, Q.~Zheng, Y.~Zhao, X.~Xu, and C.~Li, ``Dlformer: Discrete latent
  transformer for video inpainting,'' in \emph{Proc. CVPR}, 2022, pp.
  3511--3520.

\bibitem{carion2020end}
N.~Carion, F.~Massa, G.~Synnaeve, N.~Usunier, A.~Kirillov, and S.~Zagoruyko,
  ``End-to-end object detection with transformers,'' in \emph{Proc.
  ECCV}.\hskip 1em plus 0.5em minus 0.4em\relax Springer, 2020, pp. 213--229.

\bibitem{zhu2020deformable}
X.~Zhu, W.~Su, L.~Lu, B.~Li, X.~Wang, and J.~Dai, ``Deformable detr: Deformable
  transformers for end-to-end object detection,'' in \emph{ICLR}, 2021.

\bibitem{zeng2021improving}
Y.~Zeng, H.~Yang, H.~Chao, J.~Wang, and J.~Fu, ``Improving visual quality of
  image synthesis by a token-based generator with transformers,'' in
  \emph{Proc. NeurIPS}, 2021, pp. 21\,125--21\,137.

\bibitem{guo2021image}
Z.~Guo, D.~Guo, H.~Zheng, Z.~Gu, B.~Zheng, and J.~Dong, ``Image harmonization
  with transformer,'' in \emph{Proc. ICCV}, 2021, pp. 14\,870--14\,879.

\bibitem{li2020learning}
J.~Li, Y.~Yin, H.~Chu, Y.~Zhou, T.~Wang, S.~Fidler, and H.~Li, ``Learning to
  generate diverse dance motions with transformer,'' \emph{arXiv preprint
  arXiv:2008.08171}, 2020.

\bibitem{yu2020multimodal}
L.~Yu, J.~Yu, M.~Li, and Q.~Ling, ``Multimodal inputs driven talking face
  generation with spatial--temporal dependency,'' \emph{TCSVT}, vol.~31, no.~1,
  pp. 203--216, 2020.

\bibitem{lai2018learning}
W.-S. Lai, J.-B. Huang, O.~Wang, E.~Shechtman, E.~Yumer, and M.-H. Yang,
  ``Learning blind video temporal consistency,'' in \emph{Proc. ECCV}, 2018,
  pp. 170--185.

\bibitem{lei2020blind}
C.~Lei, Y.~Xing, and Q.~Chen, ``Blind video temporal consistency via deep video
  prior,'' in \emph{Proc. NeurIPS}, 2020, pp. 1083--1093.

\bibitem{matsushita2006full}
Y.~Matsushita, E.~Ofek, W.~Ge, X.~Tang, and H.-Y. Shum, ``Full-frame video
  stabilization with motion inpainting,'' \emph{TPAMI}, vol.~28, no.~7, pp.
  1150--1163, 2006.

\bibitem{liu2009content}
F.~Liu, M.~Gleicher, H.~Jin, and A.~Agarwala, ``Content-preserving warps for 3d
  video stabilization,'' \emph{ACM TOG}, vol.~28, no.~3, pp. 1--9, 2009.

\bibitem{wang2018deep}
M.~Wang, G.-Y. Yang, J.-K. Lin, S.-H. Zhang, A.~Shamir, S.-P. Lu, and S.-M. Hu,
  ``Deep online video stabilization with multi-grid warping transformation
  learning,'' \emph{TIP}, vol.~28, no.~5, pp. 2283--2292, 2018.

\bibitem{li2022deep}
C.~Li, L.~Song, S.~Chen, R.~Xie, and W.~Zhang, ``Deep online video
  stabilization using imu sensors,'' \emph{TMM}, 2022.

\bibitem{amodei2016deep}
D.~Amodei, S.~Ananthanarayanan, R.~Anubhai, J.~Bai, E.~Battenberg, C.~Case,
  J.~Casper, B.~Catanzaro, Q.~Cheng, G.~Chen \emph{et~al.}, ``Deep speech 2:
  End-to-end speech recognition in english and mandarin,'' in \emph{Proc.
  ICML}.\hskip 1em plus 0.5em minus 0.4em\relax PMLR, 2016, pp. 173--182.

\bibitem{he2016deep}
K.~He, X.~Zhang, S.~Ren, and J.~Sun, ``Deep residual learning for image
  recognition,'' in \emph{Proc. CVPR}, 2016, pp. 770--778.

\bibitem{fleet2006optical}
D.~Fleet and Y.~Weiss, ``Optical flow estimation,'' in \emph{Handbook of
  mathematical models in computer vision}.\hskip 1em plus 0.5em minus
  0.4em\relax Springer, 2006, pp. 237--257.

\bibitem{wood2021fake}
E.~Wood, T.~Baltru{\v{s}}aitis, C.~Hewitt, S.~Dziadzio, T.~J. Cashman, and
  J.~Shotton, ``Fake it till you make it: Face analysis in the wild using
  synthetic data alone,'' in \emph{Proc. ICCV}, 2021, pp. 3681--3691.

\bibitem{johnson2016perceptual}
J.~Johnson, A.~Alahi, and L.~Fei-Fei, ``Perceptual losses for real-time style
  transfer and super-resolution,'' in \emph{Proc. ECCV}.\hskip 1em plus 0.5em
  minus 0.4em\relax Springer, 2016, pp. 694--711.

\bibitem{cooke2006audio}
M.~Cooke, J.~Barker, S.~Cunningham, and X.~Shao, ``An audio-visual corpus for
  speech perception and automatic speech recognition,'' \emph{The Journal of
  the Acoustical Society of America}, vol. 120, no.~5, pp. 2421--2424, 2006.

\bibitem{assael2016lipnet}
\BIBentryALTinterwordspacing
Y.~M. Assael, B.~Shillingford, S.~Whiteson, and N.~de~Freitas, ``Lipnet:
  End-to-end sentence-level lipreading,'' \emph{GPU Technology Conference},
  2017. [Online]. Available: \url{https://github.com/Fengdalu/LipNet-PyTorch}
\BIBentrySTDinterwordspacing

\bibitem{narvekar2009no}
N.~D. Narvekar and L.~J. Karam, ``A no-reference perceptual image sharpness
  metric based on a cumulative probability of blur detection,'' in \emph{2009
  International Workshop on Quality of Multimedia Experience}.\hskip 1em plus
  0.5em minus 0.4em\relax IEEE, 2009, pp. 87--91.

\bibitem{chung2016out}
J.~S. Chung and A.~Zisserman, ``Out of time: automated lip sync in the wild,''
  in \emph{Proc. ACCV}.\hskip 1em plus 0.5em minus 0.4em\relax Springer, 2016,
  pp. 251--263.

\end{thebibliography}

% Biography text here without a photo.

% \begin{IEEEbiography}[{\includegraphics[width=1in,height=1.25in,clip,keepaspectratio]{fig1.png}}]{IEEE Publications Technology Team}
% In this paragraph you can place your educational, professional background and research and other interests.\end{IEEEbiography}

\end{document}